%% file: sample-manuscript.tex
\documentclass[acmsmall]{acmart}

\usepackage{amsfonts}
\usepackage{bm}
\usepackage{float}
\usepackage{graphicx}
\usepackage{multirow}
\usepackage{pifont}
\usepackage{soul}
\usepackage{xcolor}
\usepackage{xspace}

\usepackage{courier}

\usepackage{mathtools}

\usepackage[nofillcomment,ruled,vlined,linesnumbered,lined,boxed]{algorithm2e}

\newcommand{\cmark}{\ding{51}}

\newif\ifcomment

\commenttrue

\ifcomment
\definecolor{stelios_colour}{RGB}{144, 238, 144}
\newcommand{\stelios}[1]{\sethlcolor{stelios_colour}\hl{[\textbf{Stelios:} #1]}}
\definecolor{giannis_colour}{RGB}{191, 232, 255}
\newcommand{\giannis}[1]{\sethlcolor{giannis_colour}\hl{[\textbf{Giannis:} #1]}}
\else
\newcommand{\stelios}[1]{}
\newcommand{\giannis}[1]{}
\fi

\AtBeginDocument{%
  \providecommand\BibTeX{{%
    \normalfont B\kern-0.5em{\scshape i\kern-0.25em b}\kern-0.8em\TeX}}}

\setcopyright{acmcopyright}
\copyrightyear{2024}
\acmYear{2024}
\acmDOI{10.1145/3665868}

\acmJournal{TECS}
\acmVolume{23}
\acmNumber{4}
\acmArticle{60}
\acmMonth{6}

\acmPrice{}
\acmISBN{978-1-4503-XXXX-X/18/06}

\begin{document}

\title{\texttt{\textbf{CARI\MakeLowercase{n}}}\normalfont{\textbf{: Constraint-Aware and Responsive Inference on Heterogeneous Devices for Single- and Multi-DNN Workloads}}}

\author{Ioannis Panopoulos}
\email{ioannispanop@mail.ntua.gr}
\affiliation{%
  \institution{National Technical University of Athens}
  \city{Athens}
  \country{Greece}
}

\author{Stylianos I. Venieris}
\affiliation{%
  \institution{Samsung AI Center}
  \city{Cambridge}
  \country{UK}}
\email{s.venieris@samsung.com}

\author{Iakovos S. Venieris}
\email{venieris@cs.ece.ntua.gr}
\affiliation{%
  \institution{National Technical University of Athens}
  \city{Athens}
  \country{Greece}
}


\begin{abstract}
The relentless expansion of deep learning (DL) applications in recent years has prompted a pivotal shift towards on-device execution, driven by the urgent need for real-time processing, heightened privacy concerns, and reduced latency across diverse domains. This paper addresses the challenges inherent in optimising the execution of deep neural networks (DNNs) on mobile devices, with a focus on device heterogeneity, multi-DNN execution, and dynamic runtime adaptation. We introduce \texttt{CARIn}, a novel framework designed for the optimised deployment of both single- and multi-DNN applications under user-defined service-level objectives (SLOs). Leveraging an expressive multi-objective optimisation (MOO) framework and a runtime-aware sorting and search algorithm (\texttt{RASS}) as the MOO solver, \texttt{CARIn} facilitates efficient adaptation to dynamic conditions while addressing resource contention issues associated with multi-DNN execution. Notably, \texttt{RASS} generates a set of configurations, anticipating subsequent runtime adaptation, ensuring rapid, low-overhead adjustments in response to environmental fluctuations. Extensive evaluation across diverse tasks, including text classification, scene recognition, and face analysis, showcases the versatility of \texttt{CARIn} across various model architectures, such as Convolutional Neural Networks (CNNs) and Transformers, and realistic use cases. We observe a substantial enhancement in the fair treatment of the problem's objectives, reaching 1.92$\times$ when compared to single-model designs, and up to 10.69$\times$ in contrast to the state-of-the-art OODIn framework. Additionally, we achieve a significant gain of up to 4.06$\times$ over hardware-unaware designs in multi-DNN applications. Finally, our framework sustains its performance while effectively eliminating the time overhead associated with identifying the optimal design in response to environmental challenges.
\end{abstract}



\received{14 November 2023}
\received[revised]{9 April 2024}
\received[accepted]{7 May 2024}

\maketitle

\section{Introduction}
\label{sec:intro}
\input{sections/1_intro}

\section{Background \& Related Work}
\label{sec:related}
\input{sections/2_related}

\section{Overview of \texttt{CARI\MakeLowercase{n}}}
\label{sec:fw}
\input{sections/3_fw}

\section{Multi-Objective Optimisation Framework}
\label{sec:moo}
\input{sections/4_moo}

\section{Implementation}
\label{sec:runtime}
\input{sections/5_impl}

\section{Experimental Methodology}
\label{sec:setup}
\input{sections/6_setup}

\section{Results}
\label{sec:results}
\input{sections/7_results}

\section{Limitations \& Future Directions}
\label{sec:limitations}
\input{sections/8_limitations}

\section{Conclusion}
\label{sec:conclusion}
\input{sections/9_conclusion}

\begin{acks}
\begin{minipage}{0.25\textwidth}
    \includegraphics[width=\linewidth]{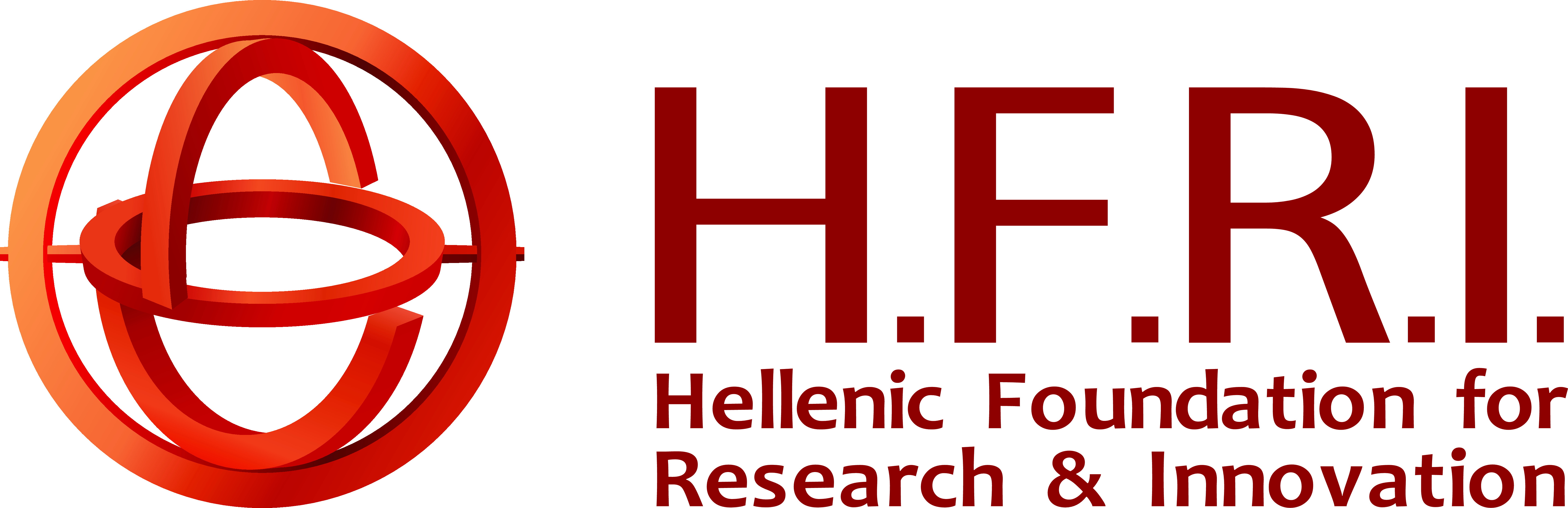}
    \Description{Funding institution.}
\end{minipage}
\hspace{0.01\textwidth} 
\begin{minipage}{0.74\textwidth}
    \vspace{0.02cm}
    This research work was supported by the Hellenic Foundation for Research and Innovation (HFRI) under the 3rd Call for HFRI PhD Fellowships (Fellowship Number: 5578).
\end{minipage}
\end{acks}

\bibliographystyle{ACM-Reference-Format}
\bibliography{sample-base}










\end{document}

%% file: sections/1_intro.tex
In recent years, the pervasive growth of deep learning (DL) applications has catalysed a paradigm shift in the field of artificial intelligence, rendering on-device execution a critical imperative~\cite{dlsmartphones2019www}. The burgeoning demand for sophisticated deep neural networks (DNNs) spans a myriad of domains, from computer vision to natural language processing, necessitating the deployment of these models directly on mobile devices. This shift from centralised to decentralised computation arises from the intrinsic requirements of real-time processing, enhanced privacy concerns, and the need for reduced latency in diverse applications. As a consequence, the optimisation of executing deep neural networks on-device has emerged as a paramount research frontier. 

While the shift towards on-device execution of DNNs represents a pivotal advancement, it is not without its formidable challenges. Device heterogeneity, characterised by the diverse array of hardware and computational capabilities across mobile devices, remains a persistent hurdle. Moreover, emerging challenges, such as the simultaneous execution of multiple DNNs on a single device~\cite{multidnn_accel2023mc} and the need for dynamic runtime adaptation~\cite{adaptivenet2023corr} to evolving environmental conditions, add layers of complexity to the optimisation landscape. Multi-DNN execution introduces intricate dependencies and resource contention issues, necessitating sophisticated orchestration strategies. Runtime adaptation, on the other hand, mandates the development of intelligent mechanisms capable of dynamically adjusting model parameters and system configurations to optimise performance in real-time scenarios. Addressing these challenges is paramount to unlocking the full potential of on-device deep learning, as it paves the way for the seamless integration of advanced AI capabilities into the fabric of our interconnected devices.

Enhancing the work presented in~\cite{oodin2021smartcomp}, namely OODIn, this paper presents \texttt{CARIn}, a novel framework for the optimised deployment of both single- and multi-DNN applications on mobile devices. The initial work in~\cite{oodin2021smartcomp} focused primarily on presenting a new highly parametrised software architecture for DL mobile apps, optimising single-DNN applications and evaluating solely on the image classification task. In this paper, we build upon the architecture of OODIn that allows us to efficiently modify model and system parameters, and introduce two novel components in order to meet the new demands of model multi-tenancy and efficient runtime adaptability. First, we develop an expressive multi-objective optimisation (MOO) framework that allows us to capture both single- and multi-DNN workloads and to formally model the performance requirements and constraints of DL applications. Second, we present \texttt{RASS}, a runtime-aware MOO solver that enables rapid, low-overhead adaptation while sustaining high performance under dynamic conditions. Contrary to existing optimisers that yield a \textit{single} execution plan for a given device~\cite{mulayer2019eurosys,wang2019high}, our solver generates a \textit{set} of configurations to accommodate potential variations in resource availability. This eliminates the necessity to continually adjust and resolve the MOO problem whenever a runtime issue arises. Additionally, we broaden the scope of targeted tasks and further augment this by conducting a comprehensive evaluation spanning various model architectures, including Convolutional Neural Networks (CNNs) and Transformers, across a spectrum of realistic scenarios characterised by diverse performance demands.

%% file: sections/2_related.tex
\subsection{Problem Statement}
\label{sec:background:challenges}
The primary aim of a DL application is to consistently uphold its performance goals or service-level objectives (SLOs), often referred to as quality of service (QoS) targets. These SLOs encompass a multifaceted range of critical metrics, including but not confined to accuracy, latency, throughput, memory utilisation and energy consumption. Achieving and sustaining these objectives requires careful consideration of the specific demands inherent to a given application or system. It is crucial to recognise that this challenge is compounded by two primary factors: (a)~the inherent heterogeneity and (b)~the dynamic nature of mobile and embedded devices. Adding to this complexity is the increasingly prevalent use of multiple models within DL applications, which places additional demands on the already intricate ecosystem of these devices.

\subsubsection{Device Heterogeneity}
\label{sec:challenges:heterogeneity}
Compact devices involve a wide array of hardware configurations, characteristics, and capabilities, leading to a significant level of diversity~\cite{embench2019emdl,ai_benchmark2019iccvw, fb_edge2019hpca, smart_at_what_cost2021imc}. This diversity manifests not only across distinct devices, which is referred to as "inter-device heterogeneity," but also within individual devices, a concept known as "intra-device heterogeneity." Inter-device heterogeneity reflects the variations in size, processing power, memory capacity, energy efficiency, and more, across different devices. For example, a high-end smartphone will have markedly different hardware specifications than a low-cost IoT sensor node, illustrating the extent of diversity that exists across various devices. Intra-device heterogeneity, on the other hand, arises from the presence of multiple hardware components and subsystems within a device, each with its distinct attributes. In a standard smartphone setup, for instance, one may encounter a CPU, a GPU, and an NPU, all with varying clock speeds, energy consumption profiles, memory requirements, and parallelisation capabilities.
\textbf{Due to device heterogeneity, it is very challenging to design a universal DL model that performs efficiently across devices}. For example, a model meticulously crafted and optimised to run seamlessly on Google's Edge TPU may encounter performance issues and inefficiencies when deployed on a different processor, such as a conventional mobile GPU.

As a result, rather than pursuing a one-size-fits-all approach, where a single model is expected to excel universally, the focus shifts towards creating models that are fine-tuned and optimised for the unique features of each target device. These device-specific models are designed to leverage the strengths of the hardware and maximise performance, thereby addressing the inherent challenges associated with device heterogeneity.

\subsubsection{Dynamic Environment}
\label{sec:challenges:dynamicity}
In contemporary mobile and embedded computing ecosystems, the concurrent execution of multiple applications and processes is commonplace. This inherent multi-tasking characteristic introduces notable fluctuations in resource availability and workload demands, thereby rendering the acquisition of sufficient resources for performant task execution a challenging endeavour. \textbf{Due to environment dynamicity, it is very challenging for a static execution configuration to consistently satisfy the application's SLOs at any given time}. For instance, if a user runs the application outdoors on a hot day, the device's temperature may rise and thermal throttling mechanisms can be triggered, causing the CPU or GPU to reduce their clock speeds to prevent overheating~\cite{dvfs_survey2020ieeedt}, resulting in reduced throughput or execution slowdown.

Such scenarios necessitate the ability to dynamically adapt to changing conditions and varying resource availability in real time. This adaptive behaviour is vital to ensure that the application consistently maintains satisfactory performance levels despite the variability in its operational environment.

\subsubsection{Multiple DNNs}
\label{sec:challenges:multidnn}
Today's growing demand for more advanced and intelligent systems has given rise to scenarios that mandate the simultaneous utilisation of multiple models, often referred to as "multi-DNN" configurations~\cite{multidnn_accel2023mc}. This paradigm shift is mainly driven by the need to address specific tasks or solve complex problems that demand a diversified approach, benefiting from the combined expertise of multiple specialised models. Multi-DNN applications showcase a high degree of adaptability in employing multiple models, as they can be harnessed to address a singular, intricate task~\cite{heimdall2020mobicom,band2022mobisys} or several distinct, autonomous tasks~\cite{virt_weights2020mobisys,masa2021percom}. In the former scenario, models typically exhibit interdependence and may require sequential execution, while in the latter scenario, models operate independently, affording them the capability to run in parallel. \textbf{The efficient deployment of multi-DNN configurations introduces intricacies that pertain to resource allocation and load distribution.} Particularly, parallel model execution presents a notably more intricate challenge compared to sequential execution, as models compete for the device's finite resources. This concurrent operation, coupled with the simultaneous management of multiple tasks, amplifies the overall workload and poses new challenges to resource allocation and coordination.

The attainment of seamless orchestration and effective collaboration among multiple specialised models while upholding stringent performance and quality benchmarks stands as a substantial and multifaceted challenge within the domain of multi-DNN applications.

\subsection{Related Work}
\label{sec:background:related_work}
The field of on-device deep learning has witnessed significant advancements in addressing the challenges posed by device heterogeneity and environment dynamicity in both single- and multi-DNN use cases. This progress reflects a profound shift in the landscape of deep learning research, where a growing emphasis has been placed on ensuring that DL models not only function effectively but also meet specific SLOs across a spectrum of computational environments, ranging from powerful high-end devices to resource-constrained edge computing platforms. Moreover, recent efforts have explored the integration of MOO techniques to further enhance the adaptability and efficiency of on-device DL solutions.

\subsubsection{Service-Level Objectives}
\label{sec:related_work:slos}
Most of the prior works directed towards achieving SLOs have predominantly concentrated on scenarios involving multiple DNNs, primarily investigating the trade-off between accuracy and latency. Within this domain, the majority is focused on system development for edge servers~\cite{bcedge2023corr, octopus2023icsoc, slo2023iwqos, fluidbatching2023iccad, dysta2023micro}, and only a limited number of studies have been devoted to on-device execution, specifically by orchestrating multiple inference requests across heterogeneous processors~\cite{band2022mobisys, slo2021taco, heimdall2020mobicom}. In contrast, our proposed framework demonstrates the capability to accommodate a diverse array of SLOs, including but not limited to accuracy, latency, memory footprint, size, and energy consumption, tailored for both single- and multi-DNN on-device applications.

\subsubsection{Multi-Objective Optimisation}
\label{sec:related_work:moo}
MOO has been widely employed in conjunction with NAS methodologies, culminating in the development of Multi-Objective Neural Architecture Search (MONAS). This approach is particularly valuable for~(a)~designing DNNs with the goal of not solely optimising the accuracy but also considering resource consumption~\cite{flexibo2023jair, monas2019iclr, mnasnet2019cvpr}, and~(b)~for compressing pretrained models~\cite{nemo2022gecco, hmq2020eccv, moocompression2021cim}. Notably, our framework represents one of the pioneering efforts to formulate and address device-specific MOO problems to achieve specific SLOs at the system level.

\subsubsection{Device-Specific Solutions}
\label{sec:related_work:heterogeneity}
The majority of endeavours aimed at addressing device heterogeneity predominantly concentrate on the model level, \textit{i.e.}~by identifying the most fitting DL architecture tailored to a specific hardware platform. Among the prominent model-level methodologies, neural architecture search (NAS) and model scaling have a central role.

Hardware-aware NAS (HW-NAS) approaches seek to optimise DNN architectures both for high predictive accuracy and for efficient execution on a target deployment platform. Its most prominent premise is the inclusion of (a)~hardware constraints, and (b)~latency, energy and other system metrics, as objectives during the search process~\cite{hadas2023date, zhang2020cvpr, aows2020cvpr, fbnet2019cvpr}. HW-NAS usually involves performance prediction in order to guide the search algorithm. Nonetheless, estimating precise latency, memory or energy figures can be challenging, and the method's effectiveness heavily relies on the accuracy of these estimates. Such approaches can also be computationally intensive due to the need to \textit{train} and evaluate a large number of candidate architectures.

Supernet-based NAS, also known as One-shot NAS, is an approach that leverages a supernet along with weight sharing to facilitate efficient architecture search~\cite{ofa2020iclr, tofa2023edge, adaptivenet2023corr, s3nas2022tcad}. A supernet is a network containing all possible architectural choices of a given search space and it enables the exploration of diverse neural architectures while significantly reducing computational overhead. While this approach reduces the training-time computational requirements, it may not be as effective at tailoring architectures to specific hardware constraints and weight sharing may restrict fine-grained control over architectural decisions.

Lastly, model scaling involves adjusting parameters such as the depth, width and input size of a DNN to strike a balance between accuracy and efficiency~\cite{scalenet2022eccv, efficientnetv22021icml, dollar2021cvpr}. This technique is often applied along with NAS or knowledge distillation methods to also accommodate resource constraints. However, model scaling might not fully exploit the unique hardware characteristics of specific devices, potentially leading to sub-optimal performance.

\subsubsection{Runtime Adaptation}
\label{sec:related_work:runtime}
In the context of runtime adaptation, research efforts span both the model and system levels. On the model level, the primary focus revolves around the development of techniques designed to dynamically adjust the model's architecture in response to fluctuations in resource availability. These adaptive models possess the capability to modify their architecture and parameters in real time during inference, effectively responding to the evolving constraints of the computing environment. Prominent examples of such models comprise adaptive supernets~\cite{chen2023pami, opa2023infocom, adaptivenet2023corr, parry2021mlcad}, adaptive model scaling~\cite{legodnn2021mobicom, mutualnet2020eccv, slimmable2019iccv}, multi-branch networks~\cite{mistify2021nsdi}, early-exit models~\cite{hapi2020iccad,hadas2023date}, and a variety of other innovative approaches. However, crafting adaptive mechanisms that seamlessly function across a diverse range of devices can pose significant technical challenges and the adaptability of these networks may introduce certain computational overhead, potentially impacting the performance of real-time applications. At the system level, complementary methods come into play, including dynamic compression~\cite{adaspring2021imwut}, adaptive model selection~\cite{kutukcu2022tecs}, and efficient scheduling on available hardware~\cite{xu2023tmc}, among others.

\subsubsection{Multi-DNN Inference}
\label{sec:related_work:multidnn}
To facilitate multi-DNN inference, researchers have also explored solutions at both the model and system levels~\cite{multidnn_accel2023mc}. From the model perspective, the execution of multiple DNNs aligns closely with the principles of multi-task learning~\cite{zhang2022tkde, sanh2019aaai, mtz2023tmc, mlink2023pami}, a technique that trains a single model to perform multiple related tasks simultaneously. Consequently, using a single multi-task model for inference can replace the need for concurrent inferences from multiple models. At the system level, most research efforts leverage the heterogeneous processors available on devices and aim to identify the highest-performing mapping strategy. This is typically achieved through approaches that partition the model at the layer level~\cite{kim2023tc, omniboost2023dac, band2022mobisys, switchflow2021middleware} or by introducing task-level priorities~\cite{rtmdl2021sensys}. Additionally, there exists a body of research focused on multi-tenant inference systems~\cite{multitenant_gpu_survey2022arxiv,multidnnmetrics2008micro}, albeit predominantly concentrating on server-based configurations~\cite{multitenant_gpu2021iccad, gmorph2024eurosys} rather than on-device implementations~\cite{multidnn_accel2023mc}.

%% file: sections/3_fw.tex
\subsection{Proposed Solution}
\texttt{CARIn} addresses the main challenges of on-device DL inference (Section~\ref{sec:background:challenges}) in two ways. First, we introduce a novel approach of modelling DL applications, utilising a multi-objective optimisation (MOO) framework to encapsulate their characteristics (Section~\ref{sec:moo}). Given the rising number and diversity of DL applications, \texttt{CARIn} is able to analytically represent their various performance requirements and constraints, with the required expressivity to support both single- and multi-DNN scenarios. Second, to enable runtime adaptation, we introduce \texttt{RASS}, a runtime-aware MOO solver that allows for low-overhead and effective dynamic adjustment of the execution. The key principle behind \texttt{RASS}'s design is to explicitly consider during the MOO solution stage that adaptation may subsequently be required at deployment time. As such, \texttt{RASS} operates in two steps: \textit{i)}~it generates a set of alternative execution configurations with diverse trade-offs prior to deployment, and \textit{ii)}~configures the inference engine with a policy of switching among them.

Towards alleviating the impact of device heterogeneity and resource fluctuation, \texttt{CARIn} operates exclusively at the system level, bypassing the need to produce an optimal model for each target device. Model-level solutions typically include the design, exploration, training, and adaptation of a DNN's architecture to specific target devices and resource availability changes. These procedures can be cumbersome, time-consuming, and lead to complex pipelines. Instead, our framework employs a repository of pre-trained models with varying architectures and complexities. The singular requisite action in relation to the models entails the application of post-training quantisation (Section~\ref{sec:quant}).

The design of our framework was driven by the fact that satisfying SLOs depends not only on the target model but also on the specific target device, especially the processor in use. Consequently, \texttt{CARIn}'s primary objective is to determine, at any given time, the most suitable model-processor\footnote{Processor here refers also to the exact configuration of a given processor, \textit{e.g.}~threads in a CPU or precision in a GPU.} pair (or pairs) for a specified device. Internally, our MOO framework expresses this as a DL-based, device-centric problem, to effectively capture both the application's SLOs and the unique characteristics of the target device. Given the device-specific nature of our MOO formulation, a distinct optimisation problem is formed for each given device, effectively circumventing the challenge posed by device heterogeneity. Additionally, in order to facilitate real-time adaptation, \texttt{CARIn} leverages the device's intra-level heterogeneity, specifically the array of available processors, as well as the range of solutions offered by the \texttt{RASS} solver, which allow the adoption of a swift and efficient switching mechanism between execution plans. 

\subsection{Workflow}
\label{sec:fw:overview}

\begin{algorithm}
    \SetAlgoLined
    \caption{\texttt{CARIn}'s end-to-end operation.}
    \label{alg:workflow}

    \KwIn{DL task(s) \newline Target device \newline Application SLOs}
    \KwOut{Designs, $\mathcal{D}$ \newline Switching Policy, SP}

    \tcc{MOO Problem Formulation}
    $\mathcal{HW} \leftarrow$ ObtainHardwareCharacteristics(Target device) \\
    
    \eIf{app type == single-DNN}{
        $\mathcal{X} \leftarrow$ ConstructDecisionSpace(DL task, $\mathcal{HW}$)
    }
    {
        $\mathcal{X} \leftarrow$ ConstructDecisionSpace(DL tasks, $\mathcal{HW}$)
    }
    
    $f_i, g_j \leftarrow$ ExtractFunctions(Application SLOs)
    
    \tcc{Objective Function Evaluation}
    $f_i(x), g_j(x) \leftarrow$ EvaluateFunctions($\mathcal{X}, f_i, g_j$) \\

    \tcc{MOO Problem Solver}
    $\mathcal{X}' \leftarrow \{ x \, | \, g_j(x) \leq 0 \}$ \quad\tcp{apply constraints}
    $opt(x) \leftarrow$ CalculateOptimality($\mathcal{X}'$) \\
    $\mathcal{X}_\text{s}' \leftarrow$ Sort($\mathcal{X}', opt$) \\
    $\mathcal{D}$, SP $\leftarrow$ Search($\mathcal{X}_\text{s}'$) \\

    \tcc{Runtime}
    \While{true}{
        $c \leftarrow$ Analyse(s) \tcp{c indicates a change in resource availability}
        \If{c == True}{
            $d_{\text{new}} \leftarrow$ RM($\mathcal{D}$, SP, $c$)  \quad \tcp{select new design}
        }
    }
\end{algorithm}

\begin{figure}
    \centering
    \includegraphics[trim=1.3cm 8.5cm 0.5cm 7.6cm, clip,width=0.99\textwidth]{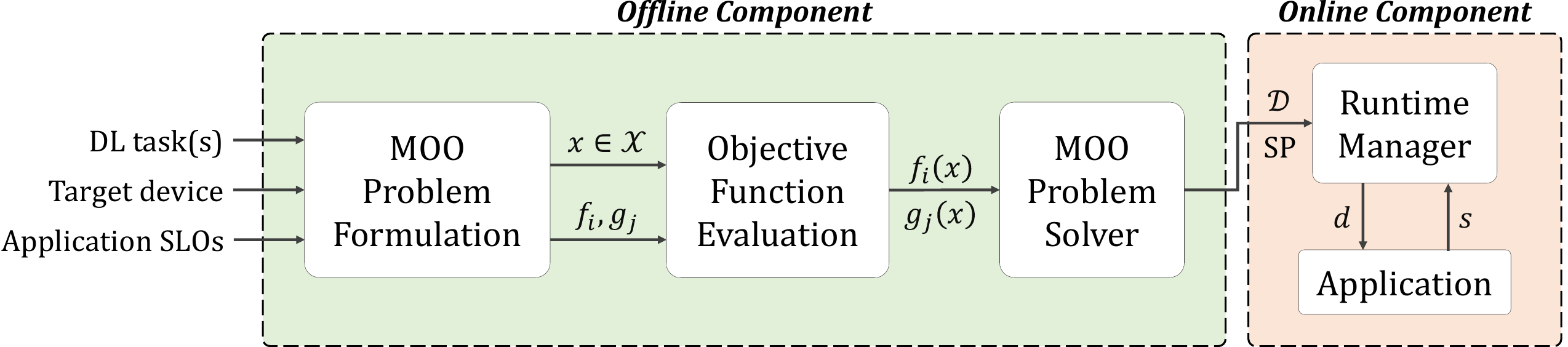}
    \caption{High-level workflow of \texttt{CARIn}.}
    \label{fig:workflow}
    \Description{Figure 1. Shows the framework's processing pipeline; inputs and outputs of the basic building blocks.}
\end{figure}

Figure~\ref{fig:workflow} depicts \texttt{CARIn}'s operational flow, which is divided into the sequential \textit{offline} and \textit{online} phases. The offline component is responsible for constructing and resolving the device-specific MOO problem. Then, at runtime, the online component's Runtime Manager (RM) constantly monitors the application's dynamic behaviour, ensuring real-time adaptation to emergent changes. Algorithm~\ref{alg:workflow} presents a comprehensive top-level overview of our framework, delineating its primary components and illustrating their main operations. These operations will be thoroughly elucidated in Section~\ref{sec:moo}. The input parameters of our framework are: (a)~the designated DL task(s) associated with the application, (b)~the stipulated SLOs, and (c)~the target device's characteristics, while the outputs consist of: (a)~the set of solutions (designs $\mathcal{D}$), and (b)~the switching policy SP.

The specified DL task or tasks dictate the set of models to be considered during the optimisation process. A model in \texttt{CARIn} is represented by the following tuple:
\begin{gather*}
    m = (arch, params, s_{\text{in}}, task, ds, pr)
\end{gather*}
where $arch$ is the model's architecture (\textit{i.e.}~layers and connections), $params$ are the model's trained parameters, $s_{\text{in}}$ is the input size, $task$ is the target DL problem, $ds$ is the name of the corresponding DL testing dataset, and $pr$ is the numerical precision to account for quantised models.

The target device defines the hardware resources at the system's disposal, which are represented by the tuple:
\begin{gather*}
    hw = (ce, op(ce))
\end{gather*}
where $ce \in \mathcal{CE}$ is the compute engine (\textit{i.e.}~processor) performing the inference computations and $op(ce)$ is a set of options tied to the given processor, \textit{e.g.}~the number of CPU threads or the GPU's numerical precision. The tuple of tunable system parameters can be extended to capture a more detailed space, \textit{e.g.}~by including the DVFS governor selection which determines the dynamic voltage and frequency scaling policy of the device~\cite{oodin2021smartcomp}. 

An individual model $m$ running under the selected system parameters $hw$ represents a single execution configuration:
\begin{equation}
    e = \left<m, hw\right> \in \mathcal{E}
    \label{eq:exec_conf}
\end{equation}

During the \textit{MOO Problem Formulation} stage (lines~1-7), \texttt{CARIn} considers every generated space of execution configurations, $\mathcal{E}_i$, in order to form the problem's decision space, $\mathcal{X}$, depending on whether the application requires single- or multi-DNN execution. At the same time, the application's SLOs delineate the MOO problem's objective functions and constraints, denoted as $f_i$ and $g_j$ respectively. Once the problem is formulated for the target device, the \textit{Objective Function Evaluation} stage evaluates each function for every $x \in \mathcal{X}$ (line~8). Following this, \texttt{CARIn}'s \textit{MOO Problem Solver} is poised to solve the MOO problem (lines~9-12). The functions CalculateOptimality, Sort, and Search shown in Algorithm~\ref{alg:workflow}, which constitute the three stages of the solver, are discussed in detail in Section~\ref{sec:moo:solver}.

In order for \texttt{CARIn} to accommodate runtime adaptation, it is important to establish a robust system for perpetually monitoring the dynamic aspects of the executing application and the state of the device itself. This ongoing vigilance enables timely recognition of abrupt alterations in operational conditions, thereby facilitating immediate corrective measures. We call this subsystem the \textit{Runtime Manager} (RM). The output of \texttt{CARIn}'s solving algorithm consists of a set $\mathcal{D}$ of highest-performing solutions, called designs, which are passed to RM along with the appropriate switching policy (SP). Leveraging a collection of periodically captured statistics $s$ from the \textit{Application}'s runtime, the RM module has the ability to discern dynamic changes in resource allocation ($c$ in Algorithm~\ref{alg:workflow}) and rapidly switch to an alternative design $d_{\text{new}}$ to effectively and robustly meet the application-level SLOs (lines~13-18).

%% file: sections/4_moo.tex
Multi-objective optimisation (MOO) constitutes a mathematical and computational approach employed to find the best solutions or trade-offs in scenarios that involve multiple interrelated and, at times, antagonistic objectives~\cite{gunantara2018cogent, pereira2022acme}. The appropriateness of a MOO framework for our problem is underscored by (a)~the inherent nature of DL application SLOs, which typically comprise objectives that exhibit conflicts, and (b)~the inherent attribute of MOO to yield a solution space rich in diversity, which, in turn, can enable dynamic adaptation.

\subsection{MOO Problem Formulation}
\label{sec:moo:formulation}
For \texttt{CARIn}'s DL-based MOO formulation, we adopt the following mathematical description:
\begin{gather*}
    \begin{aligned}
        & \text{min/max} && f_i(x) && 1 \leq i \leq N \\
        & \text{subject to} && g_j(x) = g_j(h_j(x)) \leq 0, && 1 \leq j \leq P
    \end{aligned}
\end{gather*}
where $x$ denotes the decision variable, $N$ is the number of objective functions, $f_i(x)$ is the $i$-th objective function, $P$ is the number of inequality constraints, and $g_j(x)$ is the $j$-th inequality constraint, which is always a composite function of a given inner function $h_j(x)$. Note that when there is only a single objective function ($N\!=\!1$), then the problem is reduced to single-objective optimisation (SOO). The problem's objective functions and constraints are extracted from the application's SLOs, which can be split into two categories:
\begin{itemize}
    \item Broad SLOs: Such objectives define the problem's objective functions and come in the form of {\small $\left< min/max, p \right>$}, where $p$ is a DL-related performance metric. For instance, {\small$\left<max, mIoU\right>$} means that the mean Intersection-over-Union (mIoU) accuracy metric should be maximised for an image segmentation task. For \texttt{CARIn}, this objective translates to the maximisation of the objective function $f(x) = A(x) = mIoU(x)$.
    \item Narrow SLOs: These objectives define the problem's constraints and come in the form of {\small $\left<min/max/avg/std/n^{\text{th}}, p, v \right>$}, which means that the minimum, maximum, average, standard deviation or $n$\textsuperscript{th} percentile value of $p$ is bounded by a target value $v$. For instance, {\small$\left<avg, L, 15\right>$} means that the average latency needs to be less than 15 ms, which translates to the constraint $g(x) \leq 0$, where $g(x) = g(h(x)) = g(L(x)) = \overline{L}(x) - 15$.
\end{itemize}

Given that both types of objectives concern the same set of performance metrics, it follows that both the objective and inner functions, $f_i(x)$ and $h_j(x)$, share a common function space, denoted by $\mathcal{F}$, which encompasses the entirety of available functions associated with various DL performance metrics. For this reason, in cases where the application defines constraints without explicitly specifying objective functions, \texttt{CARIn} can duly regard all specified inner functions $h_j(x)$ as objective functions as well.

\subsubsection{Single-DNN Setting}
\label{sec:moo:formulation:single}
When there is only one DL task to optimise, the decision variable $x$ is a single execution configuration $e$, as defined in Equation~\ref{eq:exec_conf}. Therefore, the execution configuration space $\mathcal{E}$ effectively transforms into the decision space $\mathcal{X}$:
\begin{gather*}
    x_{\text{single}} = e = \left<m, hw\right> \in \mathcal{X}_{\text{single}} = \mathcal{E}
\end{gather*}
For the objective functions, \texttt{CARIn} leverages the following DNN-specific performance metrics:
\begin{itemize}
    \item Size ($S$): Size is conventionally represented by either the total count of parameters within the neural network or the physical file size of the model stored in memory.
    \item Workload ($W$): This metric is typically measured in terms of numerical operations, such as floating-point operations (FLOPs) or multiply-accumulate operations (MACs).
    \item Accuracy ($A$): Accuracy is contingent upon the specific DL task in question, \textit{e.g.} top-1 accuracy for classification tasks or exact match for question answering tasks.
    \item Latency ($L$): Latency delineates the temporal lag between the transmission of input data to the DNN model and the reception of the corresponding output. It is quantified in units of milliseconds or seconds.
    \item Throughput ($TP$): Throughput provides an indication of the model's real-time processing capabilities and is computed as the total number of input samples (batch size), divided by the total inference latency. This metric is denominated in samples per second (\textit{e.g.} images per second when images constitute the inputs).
    \item Energy Consumption ($E$): This metric is of paramount importance for the evaluation of the energy efficiency of DNN applications in resource-constrained environments and is measured in energy units, such as watt-hours or joules.
    \item Memory Footprint ($MF$): Memory footprint encapsulates the extent of random-access memory (RAM) required for the loading and execution of a DNN. It is traditionally assessed in terms of memory size units, such as megabytes (MB) or gigabytes (GB).
\end{itemize}

Overall, the set of potential objective functions in single-DNN cases is denoted as:
\begin{gather*}
    \mathcal{F}_{\text{single}} = \{ S, W, A, L, TP, E, MF \}
\end{gather*}
collectively empowering a multifaceted assessment of DNN models and providing a holistic understanding of their performance across diverse dimensions.

It is important to recognise that the latency and energy consumption metrics are subject to inherent fluctuations when executing DNNs on mobile devices. These fluctuations can arise due to various factors, including device load, temperature, input values, and other environmental variables (see Section~\ref{sec:moo:solver:runtime_challenges}). As a result, relying on a single, instantaneous value may not provide a robust and representative assessment of system performance. To account for these fluctuations, \texttt{CARIn} considers statistical measures, such as the average or maximum energy consumption or the variance of the latency, as objective functions.

\subsubsection{Multi-DNN Setting}
\label{sec:moo:formulation:multi}
When there are $M$ independent DNNs to optimise jointly, the decision variable $x$ comprises of $M$ distinct execution configurations $e_i$, $1 \le i \le M$. Hence, the decision space $\mathcal{X}$ is an $M$-dimensional space, where each component of the decision variable can separately take values in the corresponding execution configuration space $\mathcal{E}_i$:
\begin{gather*}
    x_{\text{multi}} = \{ e_1, \ldots, e_M \} = \{ \left<m, hw\right>_1, \ldots, \left<m, hw\right>_M \} \in
    \mathcal{X}_{\text{multi}} = \mathcal{E}_1 \times ... \times \mathcal{E}_M
    \label{eq:input}
\end{gather*}
The array of potential objective functions is expansively broadened to encompass an additional triad of performance metrics pertaining to parallel execution~\cite{multidnnmetrics2008micro,multidnn_accel2023mc}:
\begin{itemize}
    \item Normalised Turnaround Time ($NTT$): $NTT$ serves as a quantifier of the perceived execution slowdown during multi-DNN execution. The $NTT$ for the $i$-th DNN is computed as:
    \begin{gather*}
        NTT_i = \frac{L_i^\text{M}}{L_i^\text{S}}
    \end{gather*}
    where $L_i^\text{S}$ and $L_i^\text{M}$ are the average latencies of the $i$-th DNN under the single- and multi-DNN modes. $NTT_i$ is a value greater than or equal to 1, with lower values indicating superior performance. For the sake of standardisation across models, it is common practice to calculate the average or maximum $NTT$.
    \item System Throughput ($STP$): $STP$ quantifies the accumulated single-DNN progress under multi-DNN execution and is computed as:
    \begin{gather*}
        STP = \sum_{i=1}^{M} NP_i = \sum_{i=1}^{M} \frac{1}{NTT_i} = \sum_{i=1}^{M} \frac{L_i^\text{S}}{L_i^\text{M}}
    \end{gather*}
    where $NP_i$ is the normalised progress of the $i$-th DNN. Its maximum magnitude is $M$, with higher values signifying enhanced performance.
    \item Fairness ($F$): The concept of fairness in a multi-DNN execution environment is contingent upon the equitable relative progress experienced by co-executing DNNs, in comparison to their single-DNN execution counterparts. Fairness, as denoted herein, is quantified as the minimum ratio of relative normalised progress rates observed among any two DNNs concurrently operating within the system:
    \begin{gather*}
        F = \min_{i, j} \frac{NP_i}{NP_j}
    \end{gather*}
    This metric adheres to a higher-is-better paradigm with values within the range $[0, 1]$, where 0 signifies an absence of fairness and 1 perfect fairness.
\end{itemize}

As a consequence, we augment \texttt{CARIn}'s objective function set to encompass both single-DNN metrics, which pertain to individual tasks or DNNs, and multi-DNN metrics, which characterise the collective performance of the entire system during concurrent execution:
\begin{gather*}
    \mathcal{F}_{\text{multi}} = \{ S_i, W_i, A_i, L_i, TP_i, E_i, MF_i \} \cup \{ STP, NTT, F \}, 1 \leq i \leq M
\end{gather*}

\subsection{Objective Function Evaluation}
\label{sec:moo:objective_function_eval}
Upon the formulation of the device-specific MOO problem, it becomes necessary to assess each objective function across the entire set of decision variables $x \in \mathcal{X}$. Assessing these functions is straightforward for certain objectives; however, it presents challenges for device-dependent functions like $E$ and $MF$, and those relying on latency, including $L$, $TP$, $STP$, $NTT$, and $F$. The approach adopted by \texttt{CARIn} for this evaluation involves the profiling of functions on individual target devices. In practical terms, this entails the deployment of all candidate models on each target device, followed by the measurement of each device-reliant objective function for all feasible model-processor combinations. We acknowledge that this procedure, albeit comprehensive, is inherently time-consuming and, in many instances, such as in multi-DNN cases, infeasible for seamless integration into real-world scenarios and practical applications. However, the optimisation of the evaluation process itself does not constitute a primary objective of this work. We extensively discuss potential enhancements of this aspect in Section~\ref{sec:limitations}.

\subsection{MOO Problem Solver}
\label{sec:moo:solver}
Following the formulation of the problem and the evaluation of objective functions, the conclusive stage involves resolving the optimisation problem. The initial step of the optimisation process is to apply the problem's constraints. Consequently, the decision variables are bound to the constrained decision space $\mathcal{X}'$, defined as:
\begin{gather*}
    \mathcal{X}' = \{ x\,|\, g_j(x) \leq 0, \forall j \}
\end{gather*}

MOO problems are frequently addressed using evolutionary algorithms, such as NSGA-II, SMS-EMOA, and MOEAD, or swarm-based algorithms, such as Ant Colony Optimisation (ACO) and Particle Swarm Optimisation (PSO)~\cite{sharma2022acme}. These algorithms systematically explore the decision variable space to discover the Pareto frontier, which represents the optimal trade-offs among conflicting objectives. While these algorithms excel in identifying the optimal execution configuration, we acknowledge that potential runtime issues may either alter the solution space, consequently affecting the Pareto frontier of a MOO problem, or introduce new constraints that were not considered during the problem's formulation. Consequently, to address the potential decline in performance, it becomes imperative to rerun these algorithms whenever a runtime issue arises, however, such repetitive executions are impractical for real-life applications and systems.

To address this challenge, we introduce a runtime-aware sorting and search algorithm, denoted as \texttt{RASS}, whose primary goal is to solve a device-specific MOO problem once, while concurrently addressing potential future runtime challenges. To achieve this, \texttt{RASS} considers both non-dominated and dominated solutions in a predictive manner, estimating the impact of possible runtime issues. In addition to providing the initial solution $d_0$, \texttt{RASS} also yields a set of supplementary runtime designs $d_i$, which serve as a proactive measure for runtime adaptation, \textit{i.e.} in instances where the currently employed design encounters performance issues. This approach alleviates the need for repetitive executions of optimisation algorithms.

The operation of \texttt{RASS} involves a sorting stage followed by a search stage. To accommodate both non-dominated and dominated solutions, our solving algorithm initially sorts candidate solutions according to their optimality (Section~\ref{sec:moo:solver:optimality}), a metric quantifying the distance from the problem's utopia point. Subsequently, based on this sorting, \texttt{RASS} identifies a set of solutions (Section~\ref{sec:moo:solver:designs_sp}) representing the various execution plans of the application which correspond to possible runtime issues (Section~\ref{sec:moo:solver:runtime_challenges}), along with a switching policy facilitating prompt transitioning between them (Section~\ref{sec:moo:solver:switching}) for the RM module.

\subsubsection{Optimality}
\label{sec:moo:solver:optimality} 
To quantify optimality for a given candidate solution $x \in \mathcal{X}'$, we first calculate the weighted Mahalanobis distance between the solution's objective vector, which is defined as $f(x) = [ f_{1}(x), f_{2}(x), ..., f_{n}(x) ]$, and the utopia point, represented as $up = [ up_{1}, up_{2}, ..., up_{n} ]$:
\begin{gather*}
    d(x) = \sqrt{\sum_{i=1}^{n} w^2_i \frac{\left[ f_{i}(x) - up_i \right]^2}{s^2_i} }
\end{gather*}
where $w_i$ is the user-supplied weight for the $i$-th objective, $s^2_i$ is the calculated variance of the $i$-th objective, and each component of the utopia point depends on the corresponding objective function:
\begin{gather*}
    up_{i} = \begin{cases} \max f_{i}, & \text{if } f_i \in \{A, TP, STP, F\} \\
    \min f_{i}, & \text{if } f_i \in \{S, W, L, E, MF, NTT\} \end{cases}
\end{gather*}

By utilising the Mahalanobis distance, we effectively accommodate the disparate scales of the diverse objectives. Consequently, optimality could also be regarded as a metric of fairness for the problem's objective functions. However, it is important to acknowledge that these functions may carry distinct significance for the problem, hence, we afford users the opportunity to define weights, thereby introducing a formal mechanism for enabling tailored optimisation strategies. Notably, the calculated distances range within the interval $[0, d_{\text{max}}]$, where the maximum distance is:
\begin{gather*}
    d_{\text{max}} = \sqrt{\sum_{i=1}^{n}  w^2_i \frac{\left( \max f_{i} - \min f_i \right)^2}{s^2_i} }
\end{gather*}

This factor necessitates the use of normalisation, which results in the distance being confined to the $[0, 1]$ range:
\begin{gather*}
    d_{\text{s}}(x) = \frac{d(x)}{d_{\text{max}}}
\end{gather*}

The optimality metric for each $x \in \mathcal{X}'$ can then formally be defined as the reciprocal of the scaled weighted Mahalanobis distance, thus, its range extends from $[1, +\infty)$:
\begin{gather*}
    opt(x) = \frac{1}{d_{\text{s}}(x)}
\end{gather*}
Utilising these values, the candidate solutions are sorted in descending order, resulting in the creation of the sorted decision space $\mathcal{X}_\text{s}$.

\subsubsection{Runtime Challenges}
\label{sec:moo:solver:runtime_challenges}
During the runtime of the application, a multitude of dynamic alterations in the device's resource availability may occur. These fluctuations impact our problem formulation in different ways, thus necessitating targeted approaches for their management. \texttt{CARIn} focuses on addressing two main challenges, regarding the processors and memory of the target device:
\begin{itemize}
    \item \textbf{Processor Overload or Overheating:} Processor-related concerns manifest when the processor at use is continuously subjected to sustained processing demands exceeding its peak processing capacity, primarily due to resource-intensive computational tasks. The protracted imposition of such an overload condition may subsequently lead to overheating, which means the escalation of the SoC's temperature to a critical and potentially harmful level. Overheating may also result from insufficient cooling mechanisms or other impediments hindering the effective dissipation of heat by the SoC. As a protective measure against potential harm, mobile SoCs are equipped with thermal throttling capabilities, which are activated when temperatures exceed predefined thresholds. Thermal throttling encompasses the deliberate reduction of the processor's clock speed and performance to mitigate heat generation and maintain a safe temperature range. The intricate interplay between processor overload and overheating significantly impacts performance and power consumption, underscoring the significance of diligent management and effective mitigation strategies.
    \item \textbf{Variability in RAM Utilisation:} Owing to the multifaceted nature of mobile devices, the utilisation of Random Access Memory (RAM) is also characterised by dynamic fluctuations. Within the execution scope of an application, numerous ancillary applications, processes, or services continually initiate and terminate in the background, potentially culminating in an unforeseen saturation of RAM capacity. Consequently, this phenomenon may precipitate performance-related challenges, encompassing lag, application crashes, and an overall deceleration of device functionality. Furthermore, the perpetual management of excessive RAM consumption may also entail elevated power consumption, thereby engendering consequential ramifications.
\end{itemize}

\subsubsection{Model/Processor Switching}
\label{sec:moo:solver:switching}
In response to runtime fluctuations, \texttt{CARIn}'s RM adopts a strategic approach that involves alterations to either the model (change model, CM), processor (CP), or both (CB) within the current execution plan. These three fundamental adjustments serve as effective measures for mitigating the challenges encountered during runtime. To this end, we introduce a prioritisation scheme. In the case of processor-related phenomena, \texttt{CARIn} prioritises transferring DNN execution from the currently used processors to inactive ones (CP or CB). This transition allows the overloaded or overheated processor to dissipate excess heat and gradually restore its performance. In cases where migration is not a viable option, such as in devices limited solely to CPU usage or multi-DNN scenarios where all processors are occupied, \texttt{CARIn} employs an alternative approach which involves replacing the current model with one of reduced computational workload (CM). Conversely, addressing the memory-related issue involves transitioning to a more compact model either on the same (CM) or a different processor (CB).

\subsubsection{Design Selection \& Switching Policy}
\label{sec:moo:solver:designs_sp}
A primary principle guiding the design of \texttt{RASS} is to ensure low complexity in order to facilitate rapid switching. This objective manifests in the generation of a relatively small number of designs, which in turn offers two additional distinct benefits: firstly, minimise storage requirements for the models; and secondly, maintain a concise switching policy comprising only a limited number of transition rules. For RM to determine the appropriate timing to transition to a new execution plan, several system parameters, related to (a)~the workload distribution across processors and (b)~the aggregate memory utilisation, need to be continuously monitored. These parameters are represented by the boolean variables $c_{ce}$ and $c_\text{m}$, indicating the presence of issues pertaining to a processor $ce$ and the memory, respectively.

The first step in identifying the solutions to the problem involves identifying the sets of different model-to-processor mappings viable for processor switching, \textit{i.e.} for reallocating DL execution to idle processors. Symbolising the number of these sets as $T$, in consideration of \texttt{RASS}'s need for simplicity, if $T>3$, we retain only the top three sets, corresponding to the highest attained optimality scores. Next, we partition our sorted decision space $\mathcal{X}_s$ into $T$ distinct subspaces $\mathcal{X}_i$, each corresponding to specific model-to-processor mappings and arranged in descending order of observed optimality. Regarding processor-related phenomena, we select designs associated with the highest optimality score within each set:
\begin{gather*}
    d_i = \mathcal{X}_i[0], i=0,\ldots,T-1
\end{gather*}
For the memory-related issue, we extract the solution with the smallest memory footprint:
\begin{gather*}
    d_\text{m} = \operatorname*{argmin}_{\substack{x}} MF(x), \, \, x \in \mathcal{X}_i, \, \, i=0,\ldots,T-1
\end{gather*}

Lastly, we extract complementary designs for two extreme (highly improbable) scenarios. The first one arises when all related processors present an issue, while the memory does not, prompting the extraction of the solution with the lightest workload:
\begin{gather*}
    d_\text{w} = \operatorname*{argmin}_{\substack{x}} W(x), \, \, x \in \mathcal{X}_i, \, \, i=0,\ldots,T-1
\end{gather*}
and the second scenario surfaces when both the processors and memory encounter issues simultaneously, necessitating the identification of the solution that strikes the optimal balance between memory usage and workload among $d_m$ and $d_w$:
\begin{gather*}
    d_{\text{wm}} = \begin{cases} d_\text{w}, & \text{if } C\left(MF(d_\text{w}), W(d_\text{w})\right) < C\left(MF(d_\text{m}), W(d_\text{m})\right) \\
    d_\text{m}, & \text{else} \end{cases}
\end{gather*}
where we use the normalised sum to compute the cost function $C$. Collectively, the set of designs is denoted as:
\begin{gather*}
    \mathcal{D} = \{ d_i, d_\text{m}, d_\text{w} \}, i=0, \ldots, T-1
\end{gather*}
therefore \texttt{RASS} can generate a maximum of five designs for a MOO problem, since $T\leq3$.

After establishing the set of designs, \texttt{RASS}'s final step involves crafting the rule-based switching policy, which serves as a reference for the RM module, guiding its decision-making process each time the boolean variables $c_{ce}$ or $c_\text{m}$ undergo a change in value. With the aim of ensuring simplicity and conciseness in the rule set, we ensure that the selection of a new design is contingent solely upon the state of the environmental variables and independent of the presently employed design. The rationale behind the construction of the rules is deliberately straightforward, as demonstrated in Section~\ref{sec:results:runtime_adaptation}, where two representative use cases are presented and analysed.

%% file: sections/5_impl.tex
\texttt{CARIn} is implemented in Java for the Android operating system. Its primary integration leverages the TensorFlow Lite (TFLite) package in its nightly build to facilitate on-device DNN execution, as well as its delegates to access mobile accelerators. The concurrent execution of multiple DNNs is achieved through the utilisation of the \verb|java.util.concurrent| Java package.

In order to create the model suite used for our framework's evaluation, \textit{i.e.} for model retrieval, training, and preparation, TensorFlow (v2.12.0) was employed in Python. Our framework seamlessly interfaces with TensorFlow Hub and Hugging Face model repositories, which allows researchers to easily access and experiment with a wide range of pre-trained models on publicly available datasets for their specific use cases. Additionally, the TFLite Converter's optimisation module was utilised to apply post-training quantisation to the models in order to enhance inference speed, efficiency and accelerator compatibility.

Regarding the objective function evaluation process, a diverse set of tools and libraries is employed. For accuracy assessments, we use custom evaluation scripts, as well as TFLite's image classification evaluation tool for the ImageNet ILSVRC 2012 task. Furthermore, to comprehensively capture the model's computational complexity and resource requirements, we use the \verb|tflite| Python package to count model parameters and FLOPs. Lastly, to assess the on-device performance of the models, we employ the C++-based TFLite benchmark tool\footnote{\url{https://github.com/tensorflow/tensorflow/tree/master/tensorflow/lite/tools/benchmark}}. This tool offers a comprehensive suite of measurements, encompassing execution time, memory utilisation, and other pertinent metrics, thereby providing a robust evaluation of the models' real-world performance characteristics.

%% file: sections/6_setup.tex
\begin{figure}
    \centering
    \includegraphics[trim=2.5cm 2.65cm 2.5cm 2.65cm,clip,width=0.99\textwidth]{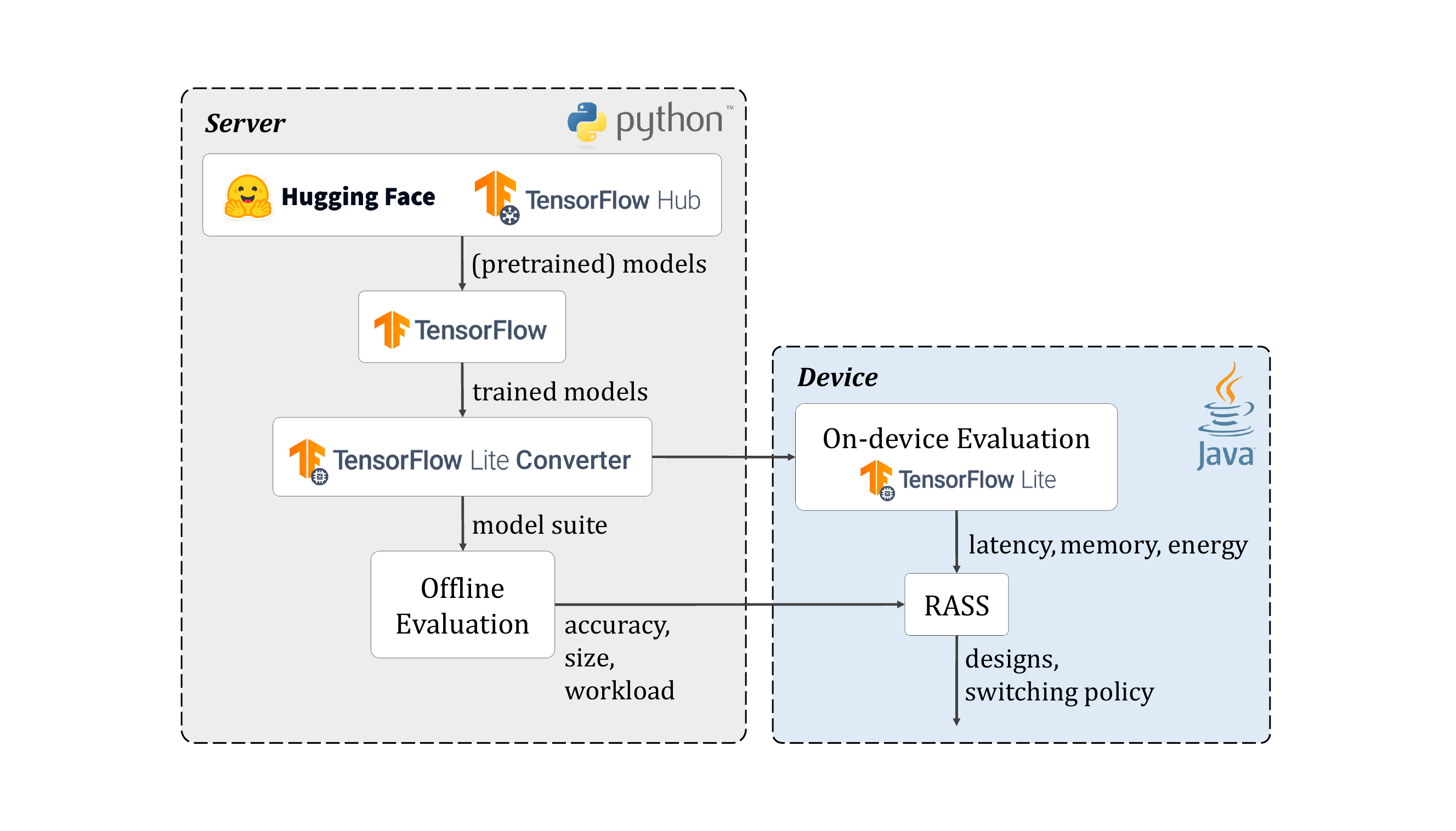}
    \caption{Toolflow for the evaluation of \texttt{CARIn}.}
    \label{fig:toolflow}
    \Description{Figure 2. Shows the tools (programming languages, frameworks, environments) and the process utilised to assess our framework.}
\end{figure}

In this section, we present the experimental methodology that underpins our research, offering insight into the comprehensive approach we have undertaken to investigate our study's core objectives. We have structured this methodology into several key subsections, each addressing a crucial aspect of our experimental design. Figure~\ref{fig:toolflow} depicts the toolflow used to conduct our experiments.

\subsection{Quantisation}
\label{sec:quant}
\texttt{CARIn} embraces \textit{post-training quantisation} as one of the most simple and mobile-friendly compression methods presently available, with benefits not only in model size, but also in latency and memory requirements. Additionally, quantisation becomes indispensable for the execution of DNNs within DSPs or NPUs designed to primarily support integer models~\cite{ai_benchmark2019iccvw}, thus unlocking complete compatibility with mobile accelerators. Notably, additional methods which also introduce trade-offs between accuracy and complexity, such as weight pruning or clustering, are orthogonal to our framework and amenable to integration. The potential synergy resulting from the combined application of various compression techniques merits further investigation.

Driven by the capabilities of the TFLite Converter, \texttt{CARIn} currently incorporates four distinct quantisation techniques, namely half-precision floating-point (FP16), 8-bit dynamic range (DR8), 8-bit fixed-point with float fallback (FX8) and full 8-bit fixed-point (FFX8). Table~\ref{tab:quant_schemes} enumerates the numerical types associated with inputs, outputs, weights, and activations for both the original (FP32) and the quantised models. It is important to note that the data type of the weights defines the storage requirements of the model. Specifically, FP16 quantisation leads to a 2$\times$ reduction in model size, while the remaining schemes (DR8, FX8, and FFX8) yield a 4$\times$ reduction in size. The operational procedures of these quantisation schemes are elucidated as follows:
\begin{itemize}
    \item FP16, by default, employs 16-bit floating-point computations, yet it possesses the flexibility to revert (fall back) to 32-bit floating-point (FP32) calculations when the hardware lacks support for 16-bit arithmetic. In such instances, the weights undergo a dequantisation process to 32-bit before the first inference. Concurrently, the activations are stored in 32-bit format. The most common processors with native support for FP16 operations are mobile GPUs.
    \item In the case of DR8, weights are represented with 8 bits, while activations persistently remain in FP32. Nevertheless, certain activations may undergo dynamic quantisation during inference, utilising quantised kernels for faster execution. The utilisation of fixed-point arithmetic, whenever feasible, may result in reduced computation times compared to relying solely on floating-point arithmetic, contingent on the specific model's characteristics.
    \item FX8, analogous to FP16, represents an 8-bit equivalent and operates with integer kernels as the default mode of execution. However, it retains the ability to utilise 32-bit operators when integer implementations are unavailable on the given hardware (floating-point fallback). Importantly, in this scheme, the converted model maintains inputs and outputs in floating-point format, allowing the model itself to determine the quantisation parameters to minimise accuracy loss.
    \item FFX8 enforces full integer quantization for all components of the model, encompassing weights, activations, operations, inputs, and outputs. This stringent quantisation scheme guarantees compatibility with integer-only devices and accelerators, such as microcontrollers, DSPs, and NPUs.
\end{itemize}

\begin{table}
  \caption{Quantisation Schemes}
  \label{tab:quant_schemes}
  \begin{tabular}{l c c c}
    \toprule
    \textbf{Scheme} & \textbf{Inputs \& Outputs} & \textbf{Weights} & \textbf{Activations} \\
    \midrule
    FP32 & fp32/int32/int64 & fp32 & fp32 \\
    FP16 & fp32/int32/int64 & fp16 & fp16/fp32 \\
    DR8 & fp32/int32/int64 & int8 & fp32 \\
    FX8 & fp32/int32/int64 & int8 & int8/fp32 \\
    FFX8 & int8/int32 & int8 & int8 \\
    \bottomrule
\end{tabular}
\end{table}

\subsection{Application Scenarios, Models and Tasks}
In the following part, we outline four discrete application use cases, which form the bedrock of our experimental evaluation. These application scenarios represent diverse real-world settings in which our research findings will be tested and validated, providing valuable insights into the effectiveness of our proposed formulation and methodology.

Notably, the first two scenarios pertain to the execution of a single DNN, while the later two involve the execution of multiple DNNs in parallel, affording us the opportunity to assess performance outcomes in instances where dependencies among multiple DNNs exist. This dichotomy is instrumental in affording a comprehensive evaluation of our methodology's versatility, applicability, and scalability. We define specific SLOs for each use case and showcase the list of models to be considered during evaluation, along with their device-independent evaluation, which includes (a)~the accuracy of both the original and quantised variants, (b)~the computational workload in FLOPs, and (c)~the model size in terms of parameters.

\subsubsection{Use Case \#1 (UC1)}
In our first single-DNN scenario, we examine the practical application of \textbf{real-time image classification}. In this setting, the camera of a mobile device continuously captures frames that require prompt and accurate recognition. The term "real-time" is qualified by a temporal restriction mandating that the maximum permissible latency is 41.67~ms, underscoring the necessity to uphold a recognition rate of no less than 24 frames per second (FPS). The principal objectives of this use case encompass the joint maximisation of accuracy and throughput. Mathematically, this MOO problem comprises two objective functions and a single constraint:
\begin{gather*}
    \begin{aligned}
        && \max &&& A(x), TP(x) \\
        && \text{subject to} &&& \text{max} \, L(x) \leq 41.67~\text{ms}
    \end{aligned}
\end{gather*}

For UC1 we used the ImageNet-1k dataset~\cite{imagenet2015ijcv}. Table~\ref{tab:uc1_models} lists the eight models under consideration, which are drawn from four distinct families: MobileNets~\cite{movilenetv22018cvpr}, EfficientNets~\cite{efficientnet2019icml}, RegNets~\cite{regnet2020cvpr}, and MobileViTs~\cite{mobilevit2022iclr}. The rationale behind this extensive model selection is to ensure a well-rounded exploration of compact and mobile-friendly architectures that span a broad spectrum, encompassing both conventional CNNs and emerging Transformer-based models. Each of these architectural paradigms exhibits unique characteristics and design principles.

It is worth noting that we also contemplated the inclusion of higher-accuracy models, such as NASNet and ConvNeXt, in our analysis. However, our assessment revealed that these models failed to meet the stipulated latency constraint. Consequently, they were excluded from our study to maintain adherence to the predefined performance criteria.

\begin{table}
  \caption{UC1 Models}
  \label{tab:uc1_models}
  \scalebox{0.76}{
  \begin{tabular}{c | l c | r | r | c c c c c}
    \toprule
    
    \multirow{2}{*}{\textbf{DL Task}} & \multirow{2}{*}{\textbf{Architecture}} & \textbf{Image} & \multirow{2}{*}{\textbf{FLOPs}} & \multirow{2}{*}{\textbf{\#Params}} & \multicolumn{5}{c}{\textbf{Top-1 Accuracy (\%)}} \\
    
    &  & \textbf{Size} &  &  & \textbf{FP32} & \textbf{FP16} & \textbf{DR8} & \textbf{FX8} & \textbf{FFX8} \\
    \midrule

    & MobileNet V2 1.0 & 224x224 & 0.60 G & 3.49 M & 71.92 & 71.96 & 71.65 & 71.28 & 71.26 \\

    & RegNetY 008 & 224x224 & 1.60 G & 6.25 M & 74.28 & 74.28 & 74.18 & 74.45 & 74.47 \\

    & MobileViT XS & 256x256 & 2.10 G & 2.31 M & 74.61 & 74.61 & - & - & - \\
    
    \textbf{Image Classification} & EfficientNet Lite0 & 224x224 & 0.77 G & 4.63 M & 75.19 & 75.23 & 75.14 & 75.09 & 75.11 \\
    on ImageNet-1k & MobileNet V2 1.4 & 224x224 & 1.16 G & 6.09 M & 75.66 & 75.68 & 75.47 & 75.41 & 75.45 \\

    & RegNetY 016 & 224x224 & 3.23 G & 11.18 M & 76.76 & 76.76 & 76.62 & 76.92 & 76.84 \\
    
    & MobileViT S & 256x256 & 4.06 G & 5.57 M & 78.31 & 78.30 & - & - & - \\
    
    & EfficientNet Lite4 & 300x300 & 5.11 G & 12.95 M & 80.81 & 80.80 & 80.78 & 80.69 & 80.71 \\

    \bottomrule
\end{tabular}
}
\end{table}

\subsubsection{Use Case \#2 (UC2)}
In our second single-DNN scenario, we study the task of \textbf{text classification}, with a particular emphasis on the memory requirements of the models. To this end, we impose a memory constraint, stipulating that the executing DNN's maximum memory footprint must not exceed 90~MB. The objectives of this use case revolve around three critical factors: minimising the average latency, reducing the model size, and maximising accuracy. Mathematically, this MOO problem encompasses three objective functions and a singular constraint:
\begin{gather*}
    \begin{aligned}
        && \min &&& \overline{L}(x), S(x) \\
        && \max &&& A(x) \\
        && \text{subject to} &&& MF(x) \leq 90~\text{MB}
    \end{aligned}
\end{gather*}

For UC2, we obtained three pre-trained Transformer models on various large datasets, including Reddit comments and 2ORC citation pairs, and subsequently fine-tuned them on Emotions~\cite{carer2018emnlp}, a dataset comprising of English Twitter messages which is employed for the task of classifying input sequences into six distinct emotions. We adopted the dataset's split configuration, which allocated 16k samples for training, 2k for validation, and 2k for testing. The reported top-1 accuracy corresponds to the dataset's test set. The selected models, detailed in Table~\ref{tab:uc2_models}, encompass the traditional BERT architecture in a lightweight version, alongside two mobile-grade models: XtremeDistil~\cite{xtremedistiltransformers2021arxiv} and MobileBERT~\cite{mobilebert2020acl}. The letter "L" in each model's name stands for the number of Transformer layers and "H" stands for the hidden dimension. In preparation for training, we further optimised BERT and XtremeDistil to enhance mobile-friendliness by replacing the GELU activation function with ReLU and substituting Layer Normalisation with Batch Normalisation~\cite{mobile_nlp_transformers2023distintsys}.

\begin{table}
  \caption{UC2 Models}
  \label{tab:uc2_models}
  \scalebox{0.76}{
  \begin{tabular}{c | l c | r | r | c c c c c}
    \toprule
    
    \multirow{2}{*}{\textbf{DL Task}} & \multirow{2}{*}{\textbf{Architecture}} & \textbf{Sequence} & \multirow{2}{*}{\textbf{FLOPs}} & \multirow{2}{*}{\textbf{\#Params}} & \multicolumn{5}{c}{\textbf{Top-1 Accuracy (\%)}} \\
    
    &  & \textbf{Length} &  &  & \textbf{FP32} & \textbf{FP16} & \textbf{DR8} & \textbf{FX8} & \textbf{FFX8} \\
    \midrule

    \multirow{2}{*}{\textbf{Text Classification}} & BERT-L2-H128   & 64  & 0.05 G  & 4.31 M   & 92.10 & 92.10 & 91.90 & 91.75 & 91.75 \\
    \multirow{2}{*}{on Emotions} & XtremeDistil-L6-H256 & 64 & 0.63 G  & 12.57 M  & 93.30 & 93.30 & 93.20 & 93.15 & 93.20 \\
    & MobileBERT-L24-H512 & 64 & 2.66 G  & 24.33 M  & 93.80 & 93.80 & 93.80 & 93.65 & 94.10 \\
    
    \bottomrule
\end{tabular}
}
\end{table}

\subsubsection{Use Case \#3 (UC3)}
In our first multi-DNN scenario, we employ two DNNs for the purpose of \textbf{scene recognition}. One DNN is dedicated to processing and classifying images, while the other can process audio data in order to identify sounds from the device's surroundings. These models operate concurrently, running in parallel, and their outputs are collectively utilised to determine the specific scene within which the mobile device is situated.

In this scenario, we seek to minimise both the average latency and its standard deviation, while simultaneously maximising the attained accuracy. We impose two latency constraints for both tasks, mandating that (a)~the average latency remains consistently below 100~ms to ensure near-real-time responsiveness, and (b)~the standard deviation of latency stays below 10 ms for minimal fluctuations. The inclusion of the latency's standard deviation aims to minimise performance variability, which still constitutes a withstanding challenge for on-device inference~\cite{fb_edge2019hpca}. Mathematically, this MOO problem is formulated as:
\begin{gather*}
    \begin{aligned}
        && \min &&& \overline{L}_i(x), \sigma_{L_{i}}(x) \\
        && \max &&& A_i(x), &  i=1,2 \\
        && \text{subject to} &&& \overline{L}_i(x) \leq 100~\text{ms}, \sigma_{L_{i}}(x) \leq 10~\text{ms}
    \end{aligned}
\end{gather*}

Table~\ref{tab:uc3_models} presents the models for each task. For the vision task, we fine-tuned three EfficientNet Lite models on the MIT Indoor Scenes dataset~\cite{mitindoorscenes2009cvpr}, which includes 67 classes and 100 images per class (80 for training and 20 for testing). We report the top-1 accuracy on the test set. For the audio task, we use YAMNet, which is trained on the AudioSet dataset~\cite{audioset2017icassp} for multi-label classification. The dataset consists of 521 sound events (classes) and 18k samples. We report the mean average precision (mAP) on the validation set. YAMNet's input waveform can vary in length. In our experiments, we use the model's minimum possible length of 975 ms, which corresponds to 15600 input samples and a total workload of 0.14 GFLOPs.

\begin{table}
  \caption{UC3 Models}
  \label{tab:uc3_models}
  \scalebox{0.76}{
  \begin{tabular}{c | l c | r | r | c c c c c}
    \toprule
    
    \multirow{2}{*}{\textbf{DL Task}} & \multirow{2}{*}{\textbf{Architecture}} & \textbf{Input} & \multirow{2}{*}{\textbf{FLOPs}} & \multirow{2}{*}{\textbf{\#Params}} & \multicolumn{5}{c}{\textbf{Accuracy}} \\
    &  & \textbf{Size} &  &  & \textbf{FP32} & \textbf{FP16} & \textbf{DR8} & \textbf{FX8} & \textbf{FFX8} \\
    
    \midrule

    \multirow{2}{*}{\textbf{Scene Classification}} & EfficientNet Lite0 & 224x224 & 0.59 G & 3.44 M & 69.78 & 69.70 & 68.96 & 69.18 & 69.18 \\
    \multirow{2}{*}{on MIT Indoor Scenes} & EfficientNet Lite2 & 260x260 & 1.51 G & 4.87 M & 76.72 & 76.72 & 77.16 & 77.69 & 77.54 \\
    & EfficientNet Lite4 & 300x300 & 4.57 G & 11.76 M & 79.33 & 79.33 & 79.18 & 79.78 & 79.48 \\
    
    \midrule

    \textbf{Audio Classification} & \multirow{2}{*}{YAMNet} & \multirow{2}{*}{15600} & \multirow{2}{*}{0.14 G} & \multirow{2}{*}{3.75 M} & \multirow{2}{*}{0.3756} & \multirow{2}{*}{0.3757} & \multirow{2}{*}{0.3620} & \multirow{2}{*}{-} & \multirow{2}{*}{-} \\
    on AudioSet   & & & & & & & & & \\
    
    \bottomrule
\end{tabular}
}
\end{table}

\subsubsection{Use Case \#4 (UC4)}
In our second multi-DNN scenario, we deploy three distinct models designed for \textbf{facial attribute prediction} tasks, namely gender, age and ethnicity estimation. These models are conceptualised as the second stage of a face detection and attribute prediction pipeline, wherein they operate concurrently on the same set of input images. As such, it is imperative for these models to adhere to stringent latency constraints to ensure minimal impact on the overall pipeline. UC4's objectives revolve around the collective optimisation of five key metrics for each model, specifically average latency, standard deviation of latency, size, memory footprint and accuracy, all while adhering to a maximum latency threshold of 10~ms. Formally:
\begin{gather*}
    \begin{aligned}
        && \min &&& \overline{L}_i(x), \sigma_{L_{i}}(x), S_i(x), MF_i(x) \\
        && \max &&& A_i(x), & i=1,2,3 \\
        && \text{subject to} &&& \text{max} \, {L}_i(x) \leq 10~\text{ms}
    \end{aligned}
\end{gather*}

In UC4, the training data are sourced from the UTKFace dataset~\cite{zhifei2017cvpr}. To ensure relevance to real-time applications, the dataset is filtered to retain samples corresponding to the age range of 18-75. Consequently, the utilised dataset comprises 18.6k facial images, partitioned into training, validation, and testing sets with a ratio of 72/8/20, respectively. The employed models leverage MobileNetV2 as the backbone architecture, extracting 576 features of size 4$\times$4, which are used for predicting the outcomes across the three distinct facial attribute prediction tasks. Notably, UC4 stands as the sole task within our study that incorporates batching during inference. Specifically, the models are configured with a batch size of 4, a choice motivated by the common case where the preceding face detection component identifies multiple faces within a single image. Table~\ref{tab:uc4_models} details the attained accuracy metrics for each task on the filtered dataset's test set: binary accuracy for gender recognition, mean absolute error for age recognition, and top-1 accuracy for ethnicity recognition across 5 output classes.
\begin{table}
  \caption{UC4 Models}
  \label{tab:uc4_models}
  \scalebox{0.76}{
  \begin{tabular}{c | l c | r | r | c c c c c}
    \toprule
    
    \multirow{2}{*}{\textbf{DL Task}} & \multirow{2}{*}{\textbf{Architecture}} & \textbf{Image} & \multirow{2}{*}{\textbf{FLOPs}} & \multirow{2}{*}{\textbf{\#Params}} & \multicolumn{5}{c}{\textbf{Accuracy}} \\
    &  & \textbf{Size} &  &  & \textbf{FP32} & \textbf{FP16} & \textbf{DR8} & \textbf{FX8} & \textbf{FFX8} \\
    
    \midrule

    \textbf{Facial Attribute} & GenderNet-MNV2 & 62x62 & 0.04 G & 0.66 M & 95.12 & 94.95 & 94.90 & 94.79 & 94.90 \\

    \textbf{Prediction} & AgeNet-MNV2 & 62x62 & 0.04 G & 0.66 M & 5.976 & 5.974 & 5.964 & 5.947 & 5.923 \\

    on UTKFace & EthniNet-MNV2 & 62x62 & 0.04 G & 0.66 M & 78.17 & 78.04 & 78.55 & 79.30 & 79.14 \\
    
    \bottomrule
\end{tabular}
}
\end{table}

\subsection{Mobile Devices}
In our study, we have selected three smartphones for our evaluation: Google Pixel 7, Samsung Galaxy S20 FE, and Samsung Galaxy A71. These devices have been deliberately chosen to represent distinct categories within the modern mobile phone landscape. A71 serves as an archetype of a mid-tier device, while S20 and P7 exemplify the high-end category, showcasing state-of-the-art features and cutting-edge technology. A detailed overview of the specifications and processing capabilities of these smartphones is shown in Table~\ref{tab:devices}.

\begin{table}
  \caption{Target Devices}
  \label{tab:devices}
  \scalebox{0.9}{
  \begin{tabular}{l l l l}
    \toprule
    \textbf{Device} & \textbf{Google Pixel 7} & \textbf{Samsung Galaxy S20 FE} & \textbf{Samsung Galaxy A71} \\
    \midrule
    Launch & 2022, October & 2020, October & 2020, January \\
    SoC & Tensor G2 & Exynos 990 & Snapdragon 730 \\
    \multirow{3}{*}{CPU} & 2$\times$2.85 GHz Cortex-X1 & 2$\times$2.73 GHz Exynos M5 & \multirow{2}{*}{2$\times$2.20 GHz Kryo 470 Gold} \\
    & 2$\times$2.35 GHz Cortex-A76 & 2$\times$2.50 GHz Cortex-A76 & \multirow{2}{*}{6$\times$1.80 GHz Kryo 470 Silver} \\
    & 4$\times$1.80 GHz Cortex-A55 & 4$\times$2.00 GHz Cortex-A55 & \\
    GPU & Mali-G710 MP7 @850 MHz & Mali-G77 MP11 @800 MHz & Adreno 618 @700 MHz  \\
    NPU & Tensor Processing Unit & \cmark & Hexagon Tensor Accelerator \\
    RAM & 8 GB @3200 MHz & 6 GB @2750 MHz & 6 GB @1866 MHz \\
    TDP & 7 W & 9 W & 5 W \\
    \bottomrule
\end{tabular}
}
\end{table}

Each of the three devices is equipped with its own NPU. Concretely, P7 incorporates a custom mobile-oriented Tensor Processing Unit (TPU), S20 features the EDEN API, which grants access to the Exynos NPU for fixed-point models and specialised GPU kernels for floating-point models, and lastly, A71 hosts the Hexagon Tensor Accelerator (HTA), a dedicated compute engine for fixed-point CNNs. Additionally, it is noteworthy that among these three devices, only A71 offers access to the device's DSP for DNN inference. Therefore, we result in the following compute engine sets for each device:
$$\mathcal{CE}_{\text{P7}} = \mathcal{CE}_{\text{S20}} = \{ \text{CPU}, \text{GPU}, \text{NPU}\}$$
$$\mathcal{CE}_{\text{A71}} = \{ \text{CPU}, \text{GPU}, \text{NPU}, \text{DSP} \}$$

\subsection{Profiling Details}
In this section, we present the available configuration options for each compute engine, $op(ce)$, within the context of an execution plan's tunable hardware parameters, which are employed by \texttt{CARIn} during the profiling phase of the device-specific objective functions. In the case of CPUs, we have the capability to tune the number of threads employed for multithreading and utilise the XNNPACK delegate, which serves as a back-end for the CPU, leveraging the XNNPACK library to provide highly optimised implementations for 32- and 16-bit floating-point computations, as well as symmetrically quantised DNN operations. Since all the devices under consideration are equipped with 8 CPU cores, the set of tunable options can be defined as follows:
$$op(\text{CPU}) = \{ N_{\text{threads}}, \text{XNNPACK} \}$$
where $N_{\text{threads}} = \{1, 2, 4, 8\}$ and $\text{XNNPACK} = \{\text{TRUE}, \text{FALSE}\}$, resulting in 8 distinct CPU execution combinations. On the other hand, for GPUs and NPUs, \texttt{CARIn} exclusively employs fp16 arithmetic when feasible, as it offers reduced latency without compromising accuracy:
$$op(\text{GPU}) = op(\text{NPU}) = \{ \text{precision = fp16} \}$$

Lastly, it should be noted that the DSP does not expose any configurable parameters, and thus its set of options can be defined as an empty set:
$$op(\text{DSP}) = \{ \}$$

In terms of the profiling process, we initiate each execution configuration with 5 warm-up runs to stabilise the target processor's performance and reduce variability. Subsequently, to gather statistically significant latency and energy consumption values, we execute each experiment 100~times. Lastly, to maintain consistent device temperatures and mitigate the risk of overheating, we incorporate a device idle period of 2 minutes prior to commencing the next set of runs.

%% file: sections/7_results.tex
This section presents the outcomes of our comprehensive evaluation of \texttt{CARIn}. Our findings provide valuable insights into the effectiveness of our framework in mitigating the challenges stemming from device heterogeneity and runtime fluctuations across both single- and multi-DNN scenarios, while concurrently meeting predetermined SLOs.

\subsection{Designs}
Our initial assessment focuses on evaluating the performance of \texttt{CARIn}'s designs within each available state, \textit{i.e.} single processor or combinations of processors for single- and multi-DNN applications respectively.

\subsubsection{Comparison Methods}
To comprehensively evaluate \texttt{CARIn}'s performance against existing methodologies, we employ three simple and empirical baselines and additionally compare against our earlier work, OODIn. Our baseline methods are formulated upon empirical observations to offer a fundamental level of performance and aid in setting a minimum performance expectation in real-world applications.
\begin{itemize}
    \item \textit{Single-architecture baseline:} The effectiveness of \texttt{CARIn} is contrasted with the traditional approach of considering a single model architecture, even if it is also accompanied by its quantised versions. This paradigm typically revolves around the selection of the model with the highest accuracy, optimal memory efficiency, compact size, and other relevant criteria.
    \item \textit{Transferred baseline:} To assess the extent to which \texttt{CARIn} addresses device heterogeneity, we utilise the transferred baseline, where the MOO problem is solved on a specific device, and the resultant designs are then applied to different devices. This baseline, being device-agnostic, overlooks the inherent characteristics and limitations of individual devices.
    \item \textit{Multi-DNN-unaware baseline:} The third baseline assesses the efficacy of our framework in handling concurrent model executions, particularly its capability to generate optimal model-to-processor mappings for multi-DNN workloads. The multi-DNN-unaware baseline dissects a multi-DNN MOO problem into $M$ single-DNN uncorrelated problems, solves each one independently and then combines the solutions.
    \item \textit{OODIn~\cite{oodin2021smartcomp}:} In our prior research, we utilised the weighted sum method as a means to address MOO problems. More precisely, OODIn aims to maximise the weighted sum obtained from the normalised objective functions. This approach fails to account for the inherent scale discrepancies among the diverse objective functions, particularly evident in DL metrics. While the utilisation of assigned weights may potentially mitigate this limitation, it necessitates prior knowledge of the statistical characteristics of the functions involved. When dealing with multi-DNN configurations, OODIn would operate as the multi-DNN-unaware baseline presented above, differing only in its utilisation of the weighted sum method instead of computing optimalities.
\end{itemize}

\begin{figure}
    \centering
    \includegraphics[width=0.99\textwidth]{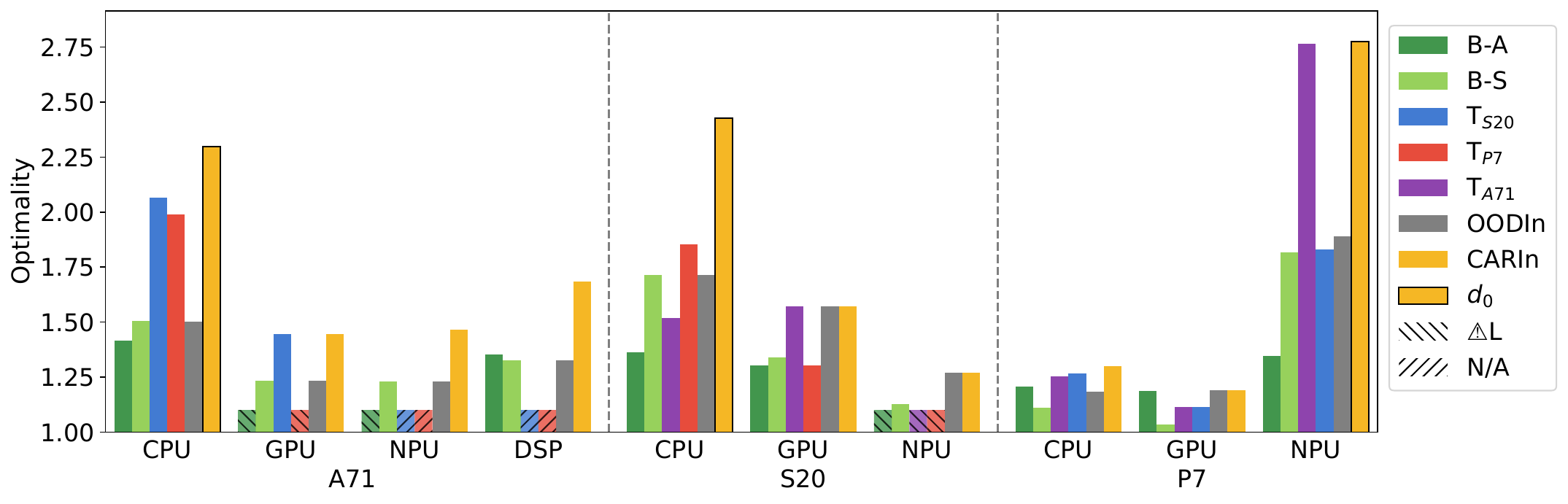}
    \caption{UC1 evaluation.}
    \label{fig:uc1_barplot}
    \Description{Figure 3. Fully described in the text.}
\end{figure}

\begin{figure}
    \centering
    \includegraphics[width=0.99\textwidth]{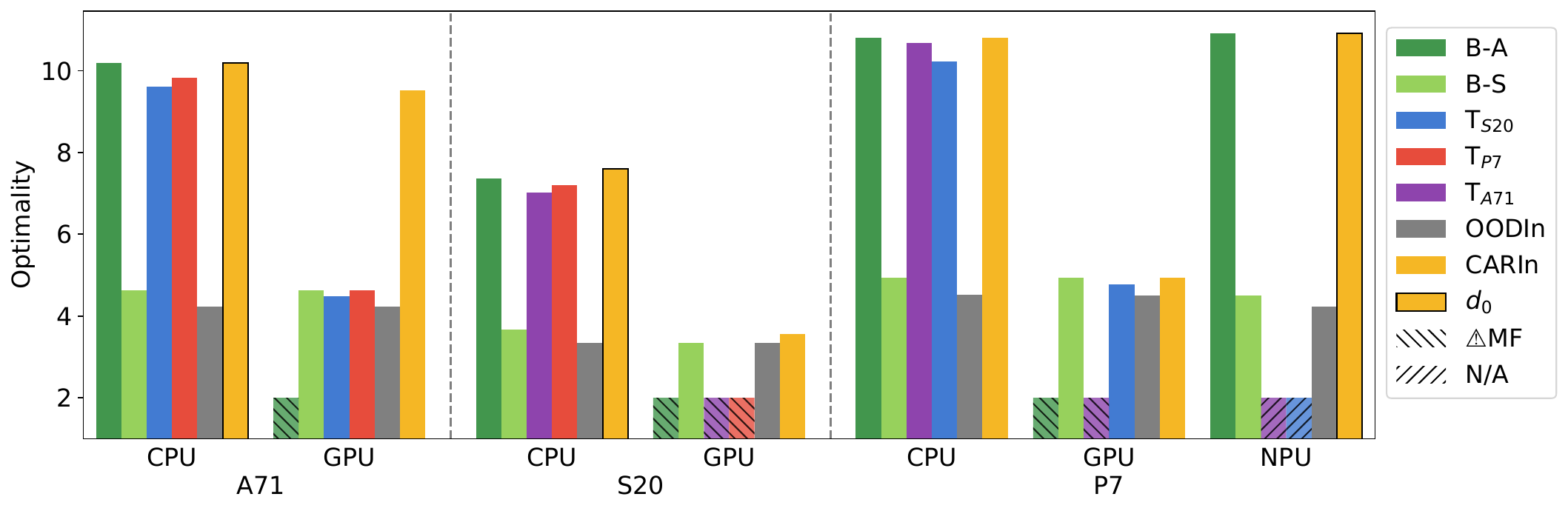}
    \caption{UC2 evaluation.}
    \label{fig:uc2_barplot}
    \Description{Figure 4. Fully described in the text.}
\end{figure}

\subsubsection{Single-DNN Execution}
Figures~\ref{fig:uc1_barplot} and ~\ref{fig:uc2_barplot} delineate the benefits of \texttt{CARIn} in relation to the optimality metric for the two single-DNN use cases. We compare against two single-architecture baselines, specifically using the model with the highest accuracy (best accuracy, B-A) and the model with the smallest size (best size, B-S), the transferred baselines from the other two devices, collectively designated as $T_{A71}$, $T_{S20}$ and $T_{P7}$, and OODIn. The initial designs $d_0$ for each device are prominently indicated, affirming the presence of device heterogeneity. Patterned bars in the figures highlight instances where certain baselines fail to yield a solution due to non-compliance with the problem's constraints (denoted by \texttt{!}) or inapplicability to different devices (denoted by \texttt{N/A}).

\textbf{Takeaways:} \textit{Our framework achieves a substantial improvement, with an average gain of 1.19$\times$ and 1.57$\times$ (up to 1.46$\times$ and 1.92$\times$) over the B-A and B-S baselines, respectively. It is noteworthy that these baselines, primarily designed for SOO problems, prove inadequate in capturing the multi-objective nature inherent in DL applications. Regarding the transferred baselines, \texttt{CARIn} achieves an average improvement of 1.17$\times$ in optimality (up to 1.84$\times$). Importantly, it not only enhances overall optimality, but also exhibits improvements across all considered objective functions. Specifically, for UC1, we observe an average increase of 0.156 units in accuracy and a 32.7\% boost in throughput, and for UC2, observable improvements include an average reduction of 2.8 MB in model size and a notable 19.9\% latency speedup at the same accuracy level. Compared to OODIn, an optimality increase of 1.5$\times$ is achieved in average (up to 1.99$\times$).}

\subsubsection{Multi-DNN Execution}
\begin{figure}
    \centering
    \includegraphics[width=0.99\textwidth]{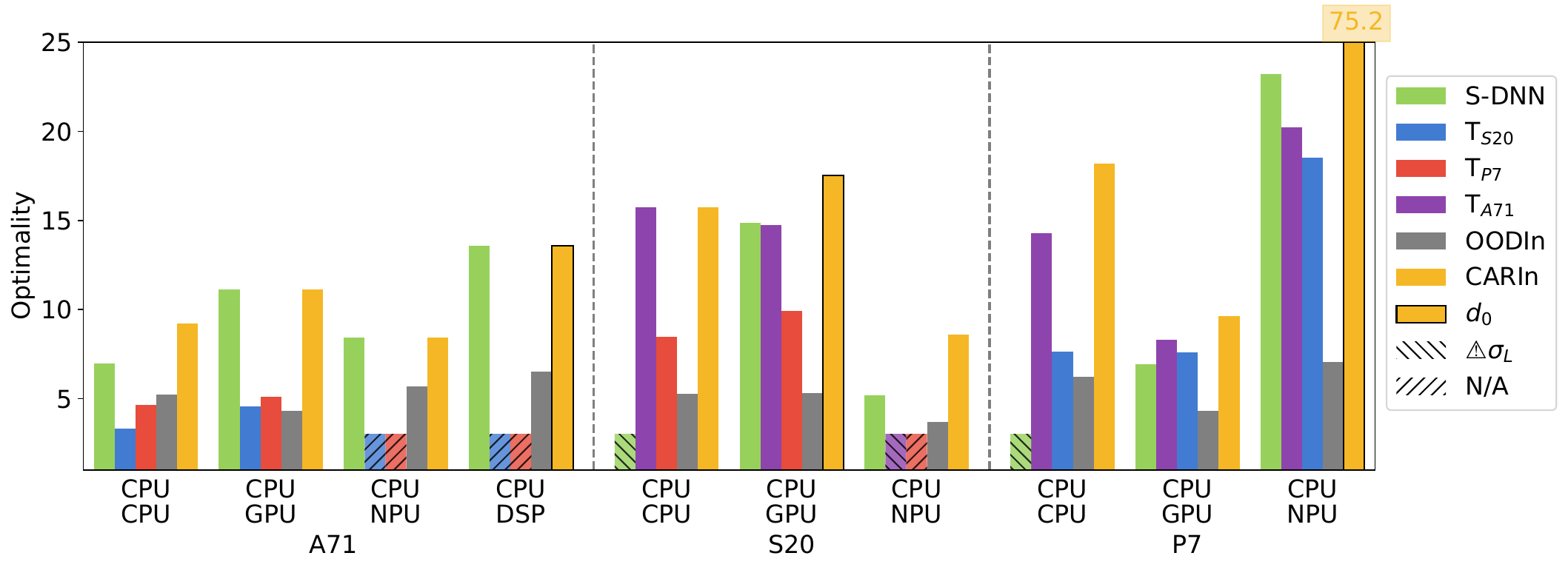}
    \caption{UC3 evaluation.}
    \label{fig:uc3_barplot}
    \Description{Figure 5. Fully described in the text.}
\end{figure}

\begin{figure}
    \centering
    \includegraphics[width=0.99\textwidth]{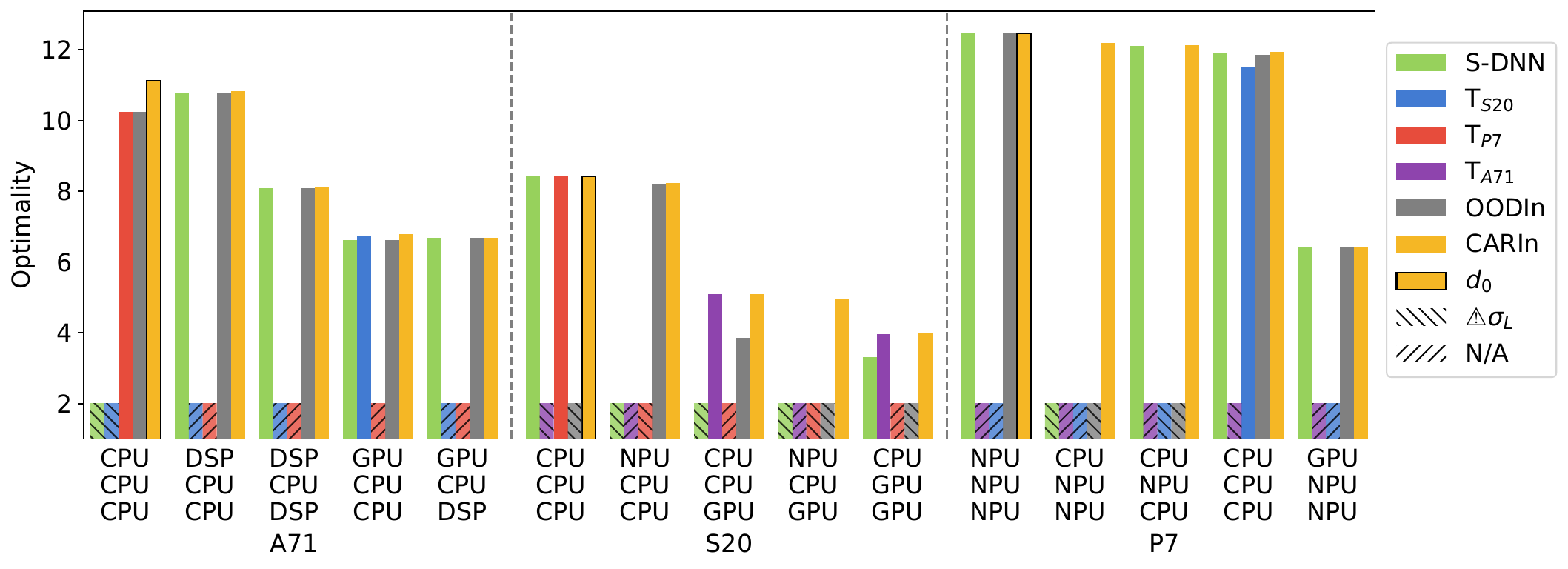}
    \caption{UC4 evaluation.}
    \label{fig:uc4_barplot}
    \Description{Figure 6. Fully described in the text.}
\end{figure}

Figures~\ref{fig:uc3_barplot} and ~\ref{fig:uc4_barplot} show the benefits of the \texttt{CARIn} framework concerning the optimality metric in the context of the two multi-DNN use cases. In these scenarios, we compare against the multi-DNN-unaware baseline, the transferred baselines from other devices, and OODIn. The horizontal axis illustrates combinations of processors for each device. In the case of UC3, all possible combinations are presented, while for UC4, due to the considerable number of combinations, we organise and display them based on optimality, showcasing the top 5 for each device.

\textbf{Takeaways:} \textit{In the context of UC3, \texttt{CARIn} delivers a significant average optimality improvement of 1.47$\times$ across devices (up to 3.24$\times$) over the multi-DNN-unaware baseline and an even more substantial gain of 1.87$\times$ (up to 4.06$\times$) over the transferred baselines. Notably, these enhancements extend across all specified objectives. Compared to OODIn, we observe a 2.83$\times$ improvement in optimality (up to 10.69$\times$). Meanwhile, UC4 poses a distinctive challenge, where the majority of baselines struggle to produce a viable solution, primarily due to their inability to satisfy the stringent latency constraints inherent in this use case, underscoring the intricacy of UC4. It is noteworthy that, given the utilisation of a singular model per task, instances where baselines do not fail result in performance parity with our framework, emphasising the importance of accommodating a diverse array of models for each task.}

\subsection{Runtime Adaptation}
\label{sec:results:runtime_adaptation}
In this section, we assess the responsiveness of the Runtime Manager (RM) and its adept utilisation of designs generated by \texttt{RASS} to dynamically adapt to a series of runtime fluctuations. For our evaluation, we target the UC1 single-DNN scenario on S20 and the UC3 multi-DNN scenario on A71. Through this experiment, we aim to demonstrate the efficacy of the RM module to seamlessly respond to dynamic runtime conditions, thereby validating its role in enhancing the adaptability and performance of \texttt{CARIn} across diverse use cases and devices.

\begin{table}
  \caption{Selected designs and switching policy for the single-DNN UC1 scenario on S20.}
  \label{tab:uc1_d_sp}
  \scalebox{0.9}{
  \begin{tabular}{c c c c | l}
    \toprule
    $\boldsymbol{c_{\text{CPU}}}$ & $\boldsymbol{c_{\text{GPU}}}$ & $\boldsymbol{c_{\text{NPU}}}$ & $\boldsymbol{c_\text{m}}$ & $\boldsymbol{d_{\text{new}}}$ \\
    \midrule
    F & - & - & F & $d_0 = \left<\text{EfficientNet Lite0 FFX8}, \text{CPU}_{4,\text{T}} \right>$  \\
    T & F & - & F & $d_1 = \left<\text{EfficientNet Lite0 FP16}, \text{GPU} \right>$   \\
    T & T & F & F & $d_2 = \left<\text{MobileNet V2 1.4 FP16}, \text{NPU} \right>$    \\
    T & T & T & F & $d_\text{w} = \left<\text{MobileNet V2 1.0 FX8}, \text{CPU}_{4,\text{T}} \right>$    \\
    T & T & T & T & $d_{\text{wm}} \equiv d_\text{w}$    \\
    - & - & - & T & $d_\text{m} = \left<\text{EfficientNet Lite0 FX8}, \text{CPU}_{8,\text{F}} \right>$  \\
    \bottomrule
\end{tabular}
}
\end{table}

\begin{figure}
    \centering
    \includegraphics[width=0.99\textwidth]{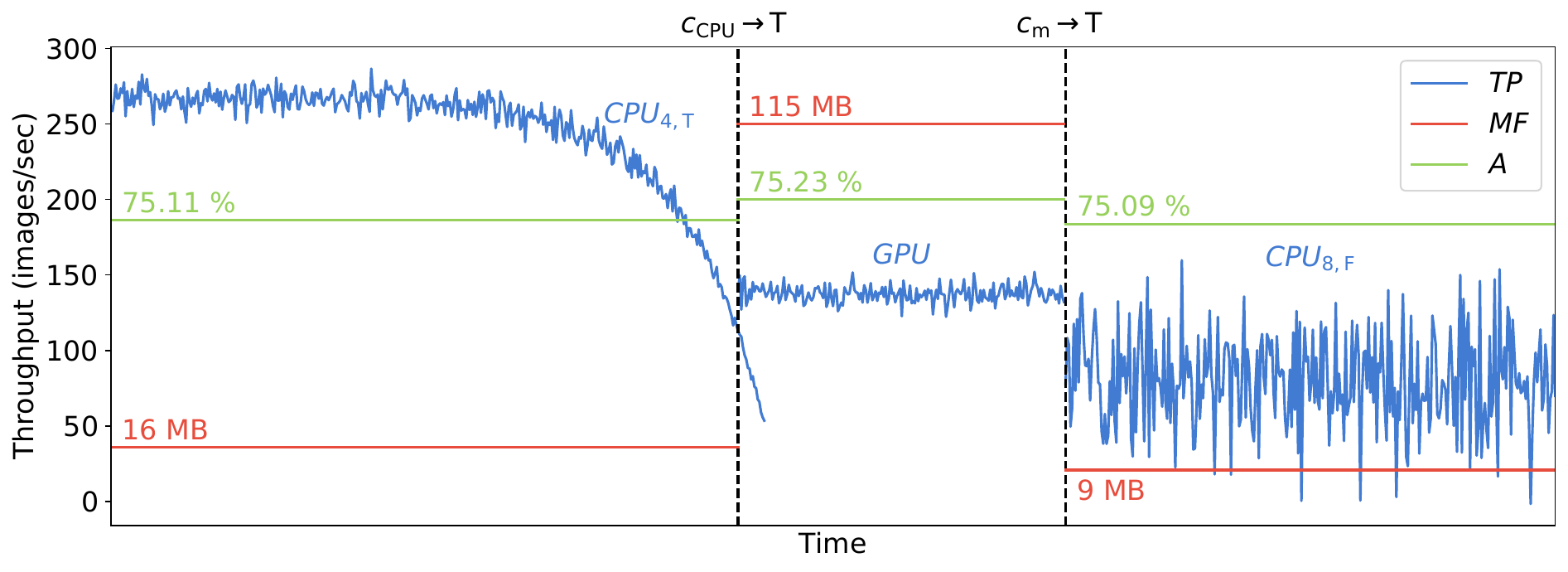}
    \caption{\texttt{CARIn}'s runtime behaviour targeting the single-DNN UC1 scenario on S20.}
    \label{fig:ra_uc1_barplot}
    \Description{Figure 7. Fully described in the text.}
\end{figure}

\subsubsection{Single-DNN Execution}
Table~\ref{tab:uc1_d_sp} presents the selected designs and switching policy, while Figure~\ref{fig:ra_uc1_barplot} depicts the behaviour of RM in the single-DNN scenario. The initial design for UC1 on S20, $d_0$, involves the utilisation of EfficientNet Lite0 FFX8 on the CPU with 4 threads and the enabled XNNPACK library, resulting in 75.11\% accuracy and a 16 MB memory footprint. As the CPU gradually becomes overloaded, the throughput experiences a decline until RM identifies an alternative design as the current highest-performing solution. The new configuration, $d_1$, entails the use of EfficientNet Lite0 FP16 on the GPU. Following further inferences, RM triggers another switch due to an impending memory issue. In this instance, \texttt{RASS} has identified the memory-efficient design, $d_\text{m}$, to involve the device's CPU.

\textbf{Takeaways:} \textit{It is worth highlighting that despite modifications in the execution plan, our framework consistently upholds accuracy levels, even when employing the memory-efficient design. This steadfast commitment to preserving user Quality of Experience (QoE) underscores \texttt{CARIn}'s resilience in the face of dynamic alterations.}

\begin{table}
  \caption{Selected designs and switching policy for the multi-DNN UC3 scenario on A71.}
  \label{tab:uc3_d_sp}
  \scalebox{0.9}{
  \begin{tabular}{c c c c | l}
    \toprule
    $\boldsymbol{c_{\text{DSP}}}$ & $\boldsymbol{c_{\text{GPU}}}$ & $\boldsymbol{c_{\text{CPU}}}$ & $\boldsymbol{c_\text{m}}$ & $\boldsymbol{d_{\text{new}}}$ \\
    \midrule
    F & - & - & F & $d_0 = \{\left<\text{YAMNet FP16}, \text{CPU}_{2,\text{F}} \right>, \left<\text{EfficientNet Lite2 FFX8}, \text{DSP} \right>\}$  \\
    T & F & - & F & $d_1 = \{\left<\text{YAMNet FP16}, \text{CPU}_{2,\text{F}} \right>, \left<\text{EfficientNet Lite2 FX8}, \text{GPU} \right>\}$   \\
    T & T & F & F & $d_2 = \{\left<\text{YAMNet FP16}, \text{CPU}_{4,\text{F}} \right>, \left<\text{EfficientNet Lite2 FFX8}, \text{CPU}_{1,\text{T}} \right>\}$    \\
    T & T & T & F & $d_\text{w} = \{\left<\text{YAMNet DR8}, \text{CPU}_{2,\text{F}} \right>, \left<\text{EfficientNet Lite0 FFX8}, \text{CPU}_{4,\text{F}} \right>\}$    \\
    T & T & T & T & $d_{\text{wm}} \equiv d_\text{w}$    \\
    - & - & - & T & $d_\text{m} \equiv d_\text{w}$  \\
    \bottomrule
\end{tabular}
}
\end{table}

\begin{figure}
    \centering
    \includegraphics[width=0.99\textwidth]{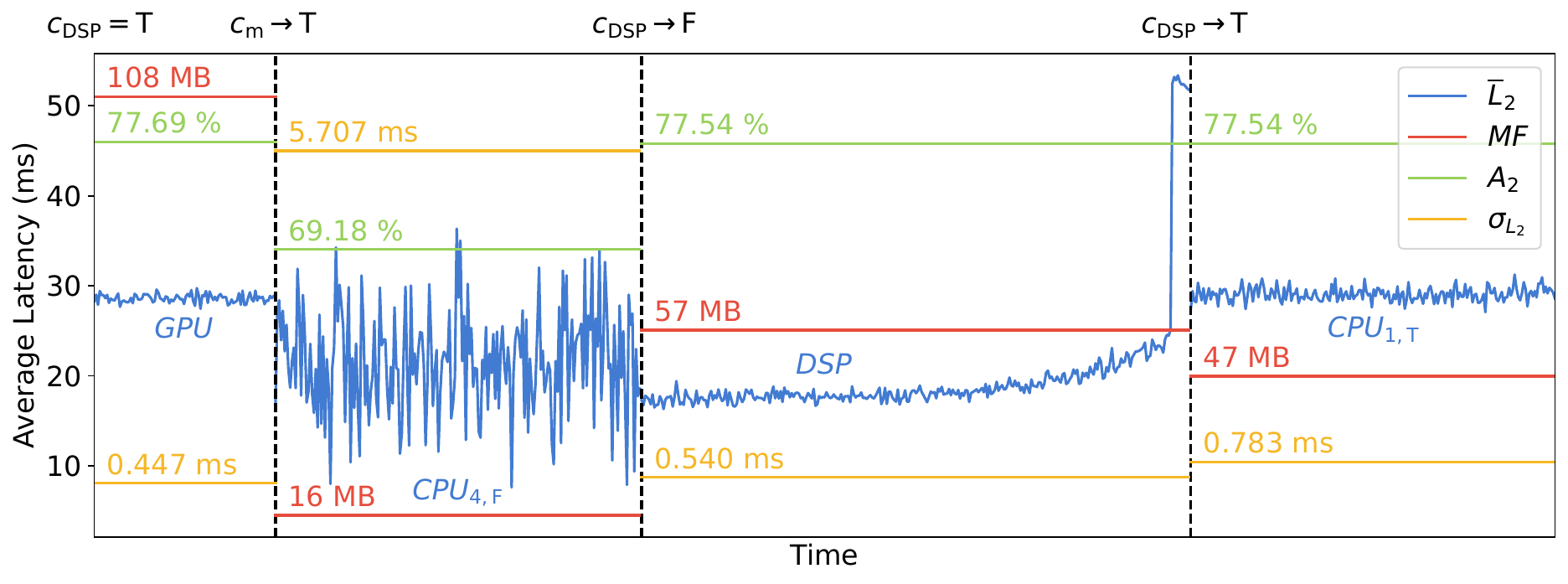}
    \caption{\texttt{CARIn}'s runtime behaviour targeting the multi-DNN UC3 scenario on A71.}
    \label{fig:ra_uc3_barplot}
    \Description{Figure 8. Fully described in the text.}
\end{figure}

\subsubsection{Multi-DNN Execution}
Table~\ref{tab:uc3_d_sp} and Figure~\ref{fig:ra_uc3_barplot} correspond to the multi-DNN scenario. In the context of UC3, where two models with distinct workloads are employed, \texttt{CARIn} recognises the heavier workload associated with the second task and acknowledges that this specific task is primarily responsible for triggering the switching mechanisms. The figure illustrates the average latency, standard deviation of latency, and accuracy for the second task, as well as the combined memory footprint of both models. UC3 involves the processing of audio data, introducing the potential use of the device's DSP for data capture and processing. Given the likelihood of DSP overload during DNN inference, suppose that the highest-performing GPU-based design, $d_1$, is currently employed with EfficientNet Lite2 FX8. However, due to the impending threat of a memory issue arising from this design's memory footprint, RM opts to switch to the memory-efficient design, $d_\text{m}$, resulting in a saving of 92 MB of RAM. Subsequently, as RM observes a reduction in DSP overload, it triggers a switch to the highest-performing design, $d_0$, characterised by lower latency and reduced memory requirements. In the event of a potential DSP overload resurgence, RM strategically avoids reverting to the GPU-based design to mitigate previous concerns of excessive memory usage. Instead, it selects the next design in line, transferring the second model to the CPU while maintaining accuracy levels.

\textbf{Takeaways:} \textit{It is important to acknowledge that \texttt{CARIn} may not always maintain predefined metric levels. As demonstrated in this instance, transitioning to the memory-efficient design resulted in an 8.5\% decrease in accuracy and an increase in jitter. However, such occurrences are considered temporary states of urgency, with a firm expectation that they will be swiftly rectified, thereby minimising impact on user QoE. Notably, the rise in average latency or the standard deviation of latency does not significantly affect user QoE, as these metrics already meet the specified latency constraints, which precede the optimisation of the objectives.}

\subsubsection{Comparison with OODIn}
In our previous work, we introduced the model/processor switching technique to mitigate runtime fluctuations. However, OODIn lacks the ability to predict forthcoming changes in resource availability, so upon detecting such events, the MOO problem necessitates readjustment to the new conditions and subsequent resolution to determine the new highest-performing solution. \texttt{CARIn} offers the advantage of solving the specified MOO problem once, prior to application initiation, thus switching to a new execution plan during runtime happens instantaneously and is based on the predetermined designs and switching policy.
Table~\ref{tab:solving_time} presents the average and maximum observed solution times of OODIn across diverse applications and devices. The solution time primarily hinges on the number of objectives and the dimensionality of the decision space $\mathcal{X}$, contingent upon the number of DL tasks, utilised models per task, compression techniques, and adjustable system parameters. Given that the time required for the TFLite interpreter to load on the CPU is typically around 3-4 ms, it becomes evident that revisiting the MOO problem can potentially become a bottleneck for the application, impacting the user's QoE.

Aside from the time overhead incurred by repeated problem solving, OODIn also requires constant access to the entire array of considered models, necessitating their storage on the user's device, which can impose limitations on the assortment of models and compression techniques initially considered. Our framework obviates the necessity to store all model variants, requiring only those selected by \texttt{RASS}. Table~\ref{tab:storage_needs} elucidates this contrast in terms of model storage requirements for every examined use case.

\begin{table}
  \caption{OODIn's solving time in milliseconds. Given that this time is inevitable whenever a runtime issue occurs, it has the potential to become a bottleneck, thus impeding the seamless execution of a DL application.}
  \label{tab:solving_time}
  \scalebox{0.9}{
  \begin{tabular}{r | r r | r r | r r}
    \toprule
    \textbf{Decision Space} & \multicolumn{2}{c|}{\textbf{A71}} & \multicolumn{2}{c|}{\textbf{S20}} & \multicolumn{2}{c}{\textbf{P7}} \\
    \textbf{Dimension} & \multicolumn{1}{c}{\textbf{\textit{Average}}} & \multicolumn{1}{c|}{\textbf{\textit{Maximum}}} & \multicolumn{1}{c}{\textbf{\textit{Average}}} & \multicolumn{1}{c|}{\textbf{\textit{Maximum}}} & \multicolumn{1}{c}{\textbf{\textit{Average}}} & \multicolumn{1}{c}{\textbf{\textit{Maximum}}} \\
    \midrule
    500 & 1.45 & 2.12 & 0.55 & 1.55 & 3.64 & 7.99 \\
    2000 & 2.80 & 5.94 & 1.70 & 3.04 & 4.94 & 9.38 \\
    5000 & 6.56 & 10.46 & 4.98 & 15.97 & 7.06 & 10.09 \\
    10000 & 12.14 & 16.07 & 11.09 & 34.25 & 10.41 & 13.38 \\
    \bottomrule
\end{tabular}
}
\end{table}

\begin{table}
  \caption{Storage requirements of \texttt{CARIn} and OODIn in MB.}
  \label{tab:storage_needs}
  \scalebox{0.9}{
  \begin{tabular}{r | r r r | r r r | r r r}
    \toprule
    & \multicolumn{3}{c|}{\textbf{A71}} & \multicolumn{3}{c|}{\textbf{S20}} & \multicolumn{3}{c}{\textbf{P7}} \\
    & \multicolumn{1}{c}{\textbf{\texttt{CARIn}}} & \multicolumn{1}{c}{\textbf{OODIn}} & \multicolumn{1}{c|}{\textbf{\textit{Reduction}}} & \multicolumn{1}{c}{\textbf{\texttt{CARIn}}} & \multicolumn{1}{c}{\textbf{OODIn}} & \multicolumn{1}{c|}{\textbf{\textit{Reduction}}} & \multicolumn{1}{c}{\textbf{\texttt{CARIn}}} & \multicolumn{1}{c}{\textbf{OODIn}} & \multicolumn{1}{c}{\textbf{\textit{Reduction}}} \\
    \midrule
    \textbf{UC1} & 13.83 & 276.36 & 19.98$\times$ & 34.37 & 443.10 & 12.89$\times$ & 34.19 & 443.10 & 12.96$\times$ \\
    \textbf{UC2} & 48.64 & 311.45 & 6.40$\times$ & 40.98 & 311.45 & 7.60$\times$ & 52.96 & 311.45 & 5.88$\times$ \\
    \textbf{UC3} & 25.74 & 205.22 & 7.97$\times$ & 58.70 & 205.22 & 3.50$\times$ & 52.81 & 205.22 & 3.89$\times$ \\
    \textbf{UC4} & 2.65 & 6.56 & 2.48$\times$ & 3.95 & 6.56 & 1.66$\times$ & 3.95 & 6.56 & 1.66$\times$ \\
    \bottomrule
\end{tabular}
}
\end{table}

%% file: sections/8_limitations.tex
In spite of the challenges mitigated by \texttt{CARIn}, our system exhibits limitations that impede its performance when deployed in practical scenarios. First, as mentioned in Section~\ref{sec:moo:objective_function_eval}, the computation of device-dependent metrics associated with objective functions or constraints across all candidate solutions is unsuitable for realistic mobile applications due to its substantial time requirements and the necessity of deploying entire models onto target devices, particularly within expansive decision spaces. Within the broader landscape of related studies, numerous works have harnessed performance prediction methodologies to estimate such metrics when executing DNNs on specific hardware platforms, without resorting to direct measurements~\cite{codl2022mobisys, herti2021pact, nnmeter2021mobisys, chamnet2019cvpr}. These models consider a range of inputs, encompassing (a)~architectural characteristics of the DNN model such as network topology, layer configurations, and overall complexity, (b)~hardware specifications including compute architecture, memory hierarchy, interconnectivity, and support for parallelism, and (c)~environmental parameters like batch size, input data characteristics, runtime conditions, and temperature/power conditions. Such approaches are orthogonal to our framework and can be integrated within \texttt{CARIn} to provide a more expedient alternative to exhaustive profiling. In the future, the exploration of such methods is envisioned to furnish a comprehensive assessment of our framework's performance and suitability for real-world scenarios.

An additional limitation arises from the selection of models for evaluation. In the contemporary landscape of generative artificial intelligence (AI)~\cite{brown2020language,touvron2023llama}, the inclusion of generative models, such as autoregressive language models, becomes paramount. These models, characterised by their ability to generate outputs sequentially based on previously generated tokens, impose heightened demands~\cite{orca2022osdi,pagedattention2023sosp}, particularly within the context of mobile environments~\cite{meltingpoint2024arxiv}. Therefore, it is imperative to account for such intricacies when assessing the efficacy of AI frameworks intended for deployment in resource-constrained settings.

%% file: sections/9_conclusion.tex
This research underscores the paramount significance of optimising the on-device execution of DNNs to meet the evolving demands of artificial intelligence applications. Building upon the foundational work of~\cite{oodin2021smartcomp}, the presented framework, \texttt{CARIn}, aims to spearhead progress in this direction. While the challenges of device heterogeneity, runtime adaptation, and multi-DNN execution persist, \texttt{CARIn} provides a novel and comprehensive solution towards alleviating them. The integration of an expressive multi-objective optimisation (MOO) framework and the introduction of \texttt{RASS} as a runtime-aware MOO solver manage to enable efficient adaptation to dynamic conditions while adhering to user-specified SLOs. \texttt{RASS} stands out for its ability to foresee upcoming runtime issues and generate a set of configurations which enable rapid, low-overhead adjustments in response to environmental fluctuations.

%% file: sample-manuscript.bbl

\begin{thebibliography}{84}


\ifx \showCODEN    \undefined \def \showCODEN     #1{\unskip}     \fi
\ifx \showDOI      \undefined \def \showDOI       #1{#1}\fi
\ifx \showISBNx    \undefined \def \showISBNx     #1{\unskip}     \fi
\ifx \showISBNxiii \undefined \def \showISBNxiii  #1{\unskip}     \fi
\ifx \showISSN     \undefined \def \showISSN      #1{\unskip}     \fi
\ifx \showLCCN     \undefined \def \showLCCN      #1{\unskip}     \fi
\ifx \shownote     \undefined \def \shownote      #1{#1}          \fi
\ifx \showarticletitle \undefined \def \showarticletitle #1{#1}   \fi
\ifx \showURL      \undefined \def \showURL       {\relax}        \fi
\providecommand\bibfield[2]{#2}
\providecommand\bibinfo[2]{#2}
\providecommand\natexlab[1]{#1}
\providecommand\showeprint[2][]{arXiv:#2}

\bibitem[Almeida et~al\mbox{.}(2019)]%
        {embench2019emdl}
\bibfield{author}{\bibinfo{person}{Mario Almeida}, \bibinfo{person}{Stefanos
  Laskaridis}, \bibinfo{person}{Ilias Leontiadis},
  \bibinfo{person}{Stylianos~I. Venieris}, {and} \bibinfo{person}{Nicholas~D.
  Lane}.} \bibinfo{year}{2019}\natexlab{}.
\newblock \showarticletitle{{EmBench: Quantifying Performance Variations of
  Deep Neural Networks Across Modern Commodity Devices}}. In
  \bibinfo{booktitle}{\emph{3rd International Workshop on Deep Learning for
  Mobile Systems and Applications (EMDL)}}.
\newblock


\bibitem[Almeida et~al\mbox{.}(2021)]%
        {smart_at_what_cost2021imc}
\bibfield{author}{\bibinfo{person}{Mario Almeida}, \bibinfo{person}{Stefanos
  Laskaridis}, \bibinfo{person}{Abhinav Mehrotra}, \bibinfo{person}{Lukasz
  Dudziak}, \bibinfo{person}{Ilias Leontiadis}, {and}
  \bibinfo{person}{Nicholas~D. Lane}.} \bibinfo{year}{2021}\natexlab{}.
\newblock \showarticletitle{{Smart at What Cost? Characterising Mobile Deep
  Neural Networks in the Wild}}. In \bibinfo{booktitle}{\emph{ACM Internet
  Measurement Conference (IMC)}}. \bibinfo{pages}{658–672}.
\newblock


\bibitem[Berman et~al\mbox{.}(2020)]%
        {aows2020cvpr}
\bibfield{author}{\bibinfo{person}{Maxim Berman}, \bibinfo{person}{Leonid
  Pishchulin}, \bibinfo{person}{Ning Xu}, \bibinfo{person}{Matthew~B.
  Blaschko}, {and} \bibinfo{person}{G{\'{e}}rard~G. Medioni}.}
  \bibinfo{year}{2020}\natexlab{}.
\newblock \showarticletitle{{AOWS:} Adaptive and Optimal Network Width Search
  With Latency Constraints}. In \bibinfo{booktitle}{\emph{2020 {IEEE/CVF}
  Conference on Computer Vision and Pattern Recognition, {CVPR} 2020, Seattle,
  WA, USA, June 13-19, 2020}}. \bibinfo{publisher}{Computer Vision Foundation /
  {IEEE}}, \bibinfo{pages}{11214--11223}.
\newblock
\urldef\tempurl%
\url{https://doi.org/10.1109/CVPR42600.2020.01123}
\showDOI{\tempurl}


\bibitem[Bouzidi et~al\mbox{.}(2023)]%
        {hadas2023date}
\bibfield{author}{\bibinfo{person}{Halima Bouzidi}, \bibinfo{person}{Mohanad
  Odema}, \bibinfo{person}{Hamza Ouarnoughi}, \bibinfo{person}{Mohammad
  Abdullah~Al Faruque}, {and} \bibinfo{person}{Sma{\"{\i}}l Niar}.}
  \bibinfo{year}{2023}\natexlab{}.
\newblock \showarticletitle{{HADAS:} Hardware-Aware Dynamic Neural Architecture
  Search for Edge Performance Scaling}. In \bibinfo{booktitle}{\emph{Design,
  Automation {\&} Test in Europe Conference {\&} Exhibition, {DATE} 2023,
  Antwerp, Belgium, April 17-19, 2023}}. \bibinfo{publisher}{{IEEE}},
  \bibinfo{pages}{1--6}.
\newblock
\urldef\tempurl%
\url{https://doi.org/10.23919/DATE56975.2023.10137095}
\showDOI{\tempurl}


\bibitem[Brown et~al\mbox{.}(2020)]%
        {brown2020language}
\bibfield{author}{\bibinfo{person}{Tom Brown}, \bibinfo{person}{Benjamin Mann},
  \bibinfo{person}{Nick Ryder}, \bibinfo{person}{Melanie Subbiah},
  \bibinfo{person}{Jared~D Kaplan}, \bibinfo{person}{Prafulla Dhariwal},
  \bibinfo{person}{Arvind Neelakantan}, \bibinfo{person}{Pranav Shyam},
  \bibinfo{person}{Girish Sastry}, \bibinfo{person}{Amanda Askell},
  {et~al\mbox{.}}} \bibinfo{year}{2020}\natexlab{}.
\newblock \showarticletitle{Language Models are Few-Shot Learners}.
\newblock \bibinfo{journal}{\emph{Advances in Neural Information Processing
  Systems (NeurIPS)}}  \bibinfo{volume}{33} (\bibinfo{year}{2020}),
  \bibinfo{pages}{1877--1901}.
\newblock


\bibitem[Cai et~al\mbox{.}(2020)]%
        {ofa2020iclr}
\bibfield{author}{\bibinfo{person}{Han Cai}, \bibinfo{person}{Chuang Gan},
  \bibinfo{person}{Tianzhe Wang}, \bibinfo{person}{Zhekai Zhang}, {and}
  \bibinfo{person}{Song Han}.} \bibinfo{year}{2020}\natexlab{}.
\newblock \showarticletitle{{Once-for-All: Train One Network and Specialize it
  for Efficient Deployment}}. In \bibinfo{booktitle}{\emph{International
  Conference on Learning Representations (ICLR)}}.
\newblock


\bibitem[Chen et~al\mbox{.}(2023)]%
        {chen2023pami}
\bibfield{author}{\bibinfo{person}{Bohong Chen}, \bibinfo{person}{Mingbao Lin},
  \bibinfo{person}{Rongrong Ji}, {and} \bibinfo{person}{Liujuan Cao}.}
  \bibinfo{year}{2023}\natexlab{}.
\newblock \showarticletitle{Prioritized Subnet Sampling for Resource-Adaptive
  Supernet Training}.
\newblock \bibinfo{journal}{\emph{{IEEE} Trans. Pattern Anal. Mach. Intell.}}
  \bibinfo{volume}{45}, \bibinfo{number}{9} (\bibinfo{year}{2023}),
  \bibinfo{pages}{11108--11119}.
\newblock
\urldef\tempurl%
\url{https://doi.org/10.1109/TPAMI.2023.3265198}
\showDOI{\tempurl}


\bibitem[Cox et~al\mbox{.}(2021)]%
        {masa2021percom}
\bibfield{author}{\bibinfo{person}{Bart Cox}, \bibinfo{person}{Jeroen
  Galjaard}, \bibinfo{person}{Amirmasoud Ghiassi}, \bibinfo{person}{Robert
  Birke}, {and} \bibinfo{person}{Lydia~Y Chen}.}
  \bibinfo{year}{2021}\natexlab{}.
\newblock \showarticletitle{{Masa: Responsive Multi-DNN Inference on the
  Edge}}. In \bibinfo{booktitle}{\emph{IEEE International Conference on
  Pervasive Computing and Communications (PerCom)}}.
\newblock


\bibitem[Dai et~al\mbox{.}(2019)]%
        {chamnet2019cvpr}
\bibfield{author}{\bibinfo{person}{Xiaoliang Dai}, \bibinfo{person}{Peizhao
  Zhang}, \bibinfo{person}{Bichen Wu}, \bibinfo{person}{Hongxu Yin},
  \bibinfo{person}{Fei Sun}, \bibinfo{person}{Yanghan Wang},
  \bibinfo{person}{Marat Dukhan}, \bibinfo{person}{Yunqing Hu},
  \bibinfo{person}{Yiming Wu}, \bibinfo{person}{Yangqing Jia},
  \bibinfo{person}{Peter Vajda}, \bibinfo{person}{Matt Uyttendaele}, {and}
  \bibinfo{person}{Niraj~K. Jha}.} \bibinfo{year}{2019}\natexlab{}.
\newblock \showarticletitle{ChamNet: Towards Efficient Network Design Through
  Platform-Aware Model Adaptation}. In \bibinfo{booktitle}{\emph{{IEEE}
  Conference on Computer Vision and Pattern Recognition, {CVPR} 2019, Long
  Beach, CA, USA, June 16-20, 2019}}. \bibinfo{publisher}{Computer Vision
  Foundation / {IEEE}}, \bibinfo{pages}{11398--11407}.
\newblock
\urldef\tempurl%
\url{https://doi.org/10.1109/CVPR.2019.01166}
\showDOI{\tempurl}


\bibitem[Doll{\'{a}}r et~al\mbox{.}(2021)]%
        {dollar2021cvpr}
\bibfield{author}{\bibinfo{person}{Piotr Doll{\'{a}}r}, \bibinfo{person}{Mannat
  Singh}, {and} \bibinfo{person}{Ross~B. Girshick}.}
  \bibinfo{year}{2021}\natexlab{}.
\newblock \showarticletitle{Fast and Accurate Model Scaling}. In
  \bibinfo{booktitle}{\emph{{IEEE} Conference on Computer Vision and Pattern
  Recognition, {CVPR} 2021, virtual, June 19-25, 2021}}.
  \bibinfo{publisher}{Computer Vision Foundation / {IEEE}},
  \bibinfo{pages}{924--932}.
\newblock
\urldef\tempurl%
\url{https://doi.org/10.1109/CVPR46437.2021.00098}
\showDOI{\tempurl}


\bibitem[Elsken et~al\mbox{.}(2019)]%
        {monas2019iclr}
\bibfield{author}{\bibinfo{person}{Thomas Elsken}, \bibinfo{person}{Jan~Hendrik
  Metzen}, {and} \bibinfo{person}{Frank Hutter}.}
  \bibinfo{year}{2019}\natexlab{}.
\newblock \showarticletitle{Efficient Multi-Objective Neural Architecture
  Search via Lamarckian Evolution}. In \bibinfo{booktitle}{\emph{7th
  International Conference on Learning Representations, {ICLR} 2019, New
  Orleans, LA, USA, May 6-9, 2019}}. \bibinfo{publisher}{OpenReview.net}.
\newblock
\urldef\tempurl%
\url{https://openreview.net/forum?id=ByME42AqK7}
\showURL{%
\tempurl}


\bibitem[Eyerman and Eeckhout(2008)]%
        {multidnnmetrics2008micro}
\bibfield{author}{\bibinfo{person}{Stijn Eyerman} {and} \bibinfo{person}{Lieven
  Eeckhout}.} \bibinfo{year}{2008}\natexlab{}.
\newblock \showarticletitle{System-Level Performance Metrics for Multiprogram
  Workloads}.
\newblock \bibinfo{journal}{\emph{{IEEE} Micro}} \bibinfo{volume}{28},
  \bibinfo{number}{3} (\bibinfo{year}{2008}), \bibinfo{pages}{42--53}.
\newblock
\urldef\tempurl%
\url{https://doi.org/10.1109/MM.2008.44}
\showDOI{\tempurl}


\bibitem[Fan et~al\mbox{.}(2023)]%
        {dysta2023micro}
\bibfield{author}{\bibinfo{person}{Hongxiang Fan}, \bibinfo{person}{},
  \bibinfo{person}{Stylianos~I. Venieris}, \bibinfo{person}{Alexandros Kouris},
  {and} \bibinfo{person}{Nicholas~D. Lane}.} \bibinfo{year}{2023}\natexlab{}.
\newblock \showarticletitle{{Sparse-DySta: Sparsity-Aware Dynamic and Static
  Scheduling for Sparse Multi-DNN Workloads}}. In
  \bibinfo{booktitle}{\emph{56th Annual IEEE/ACM International Symposium on
  Microarchitecture (MICRO)}}.
\newblock


\bibitem[Gemmeke et~al\mbox{.}(2017)]%
        {audioset2017icassp}
\bibfield{author}{\bibinfo{person}{Jort~F. Gemmeke}, \bibinfo{person}{Daniel
  P.~W. Ellis}, \bibinfo{person}{Dylan Freedman}, \bibinfo{person}{Aren
  Jansen}, \bibinfo{person}{Wade Lawrence}, \bibinfo{person}{R.~Channing
  Moore}, \bibinfo{person}{Manoj Plakal}, {and} \bibinfo{person}{Marvin
  Ritter}.} \bibinfo{year}{2017}\natexlab{}.
\newblock \showarticletitle{Audio Set: An ontology and human-labeled dataset
  for audio events}. In \bibinfo{booktitle}{\emph{2017 {IEEE} International
  Conference on Acoustics, Speech and Signal Processing, {ICASSP} 2017, New
  Orleans, LA, USA, March 5-9, 2017}}. \bibinfo{publisher}{{IEEE}},
  \bibinfo{pages}{776--780}.
\newblock
\urldef\tempurl%
\url{https://doi.org/10.1109/ICASSP.2017.7952261}
\showDOI{\tempurl}


\bibitem[Gunantara(2018)]%
        {gunantara2018cogent}
\bibfield{author}{\bibinfo{person}{Nyoman Gunantara}.}
  \bibinfo{year}{2018}\natexlab{}.
\newblock \showarticletitle{A review of multi-objective optimization: Methods
  and its applications}.
\newblock \bibinfo{journal}{\emph{Cogent Engineering}} \bibinfo{volume}{5},
  \bibinfo{number}{1} (\bibinfo{year}{2018}), \bibinfo{pages}{1502242}.
\newblock
\urldef\tempurl%
\url{https://doi.org/10.1080/23311916.2018.1502242}
\showDOI{\tempurl}
\showeprint{https://doi.org/10.1080/23311916.2018.1502242}


\bibitem[Guo et~al\mbox{.}(2023)]%
        {opa2023infocom}
\bibfield{author}{\bibinfo{person}{Junpeng Guo}, \bibinfo{person}{Shengqing
  Xia}, {and} \bibinfo{person}{Chunyi Peng}.} \bibinfo{year}{2023}\natexlab{}.
\newblock \showarticletitle{{OPA:} One-Predict-All For Efficient Deployment}.
  In \bibinfo{booktitle}{\emph{{IEEE} {INFOCOM} 2023 - {IEEE} Conference on
  Computer Communications, New York City, NY, USA, May 17-20, 2023}}.
  \bibinfo{publisher}{{IEEE}}, \bibinfo{pages}{1--10}.
\newblock
\urldef\tempurl%
\url{https://doi.org/10.1109/INFOCOM53939.2023.10228928}
\showDOI{\tempurl}


\bibitem[Guo et~al\mbox{.}(2021)]%
        {mistify2021nsdi}
\bibfield{author}{\bibinfo{person}{Peizhen Guo}, \bibinfo{person}{Bo Hu}, {and}
  \bibinfo{person}{Wenjun Hu}.} \bibinfo{year}{2021}\natexlab{}.
\newblock \showarticletitle{Mistify: Automating {DNN} Model Porting for
  On-Device Inference at the Edge}. In \bibinfo{booktitle}{\emph{18th {USENIX}
  Symposium on Networked Systems Design and Implementation, {NSDI} 2021, April
  12-14, 2021}}, \bibfield{editor}{\bibinfo{person}{James Mickens} {and}
  \bibinfo{person}{Renata Teixeira}} (Eds.). \bibinfo{publisher}{{USENIX}
  Association}, \bibinfo{pages}{705--719}.
\newblock
\urldef\tempurl%
\url{https://www.usenix.org/conference/nsdi21/presentation/guo}
\showURL{%
\tempurl}


\bibitem[Habi et~al\mbox{.}(2020)]%
        {hmq2020eccv}
\bibfield{author}{\bibinfo{person}{Hai~Victor Habi}, \bibinfo{person}{Roy~H.
  Jennings}, {and} \bibinfo{person}{Arnon Netzer}.}
  \bibinfo{year}{2020}\natexlab{}.
\newblock \showarticletitle{{HMQ:} Hardware Friendly Mixed Precision
  Quantization Block for CNNs}. In \bibinfo{booktitle}{\emph{Computer Vision -
  {ECCV} 2020 - 16th European Conference, Glasgow, UK, August 23-28, 2020,
  Proceedings, Part {XXVI}}} \emph{(\bibinfo{series}{Lecture Notes in Computer
  Science}, Vol.~\bibinfo{volume}{12371})},
  \bibfield{editor}{\bibinfo{person}{Andrea Vedaldi}, \bibinfo{person}{Horst
  Bischof}, \bibinfo{person}{Thomas Brox}, {and} \bibinfo{person}{Jan{-}Michael
  Frahm}} (Eds.). \bibinfo{publisher}{Springer}, \bibinfo{pages}{448--463}.
\newblock
\urldef\tempurl%
\url{https://doi.org/10.1007/978-3-030-58574-7\_27}
\showDOI{\tempurl}


\bibitem[Han and Baek(2021)]%
        {herti2021pact}
\bibfield{author}{\bibinfo{person}{Myeonggyun Han} {and}
  \bibinfo{person}{Woongki Baek}.} \bibinfo{year}{2021}\natexlab{}.
\newblock \showarticletitle{{HERTI:} {A} Reinforcement Learning-Augmented
  System for Efficient Real-Time Inference on Heterogeneous Embedded Systems}.
  In \bibinfo{booktitle}{\emph{30th International Conference on Parallel
  Architectures and Compilation Techniques, {PACT} 2021, Atlanta, GA, USA,
  September 26-29, 2021}}, \bibfield{editor}{\bibinfo{person}{Jaejin Lee} {and}
  \bibinfo{person}{Albert Cohen}} (Eds.). \bibinfo{publisher}{{IEEE}},
  \bibinfo{pages}{90--102}.
\newblock
\urldef\tempurl%
\url{https://doi.org/10.1109/PACT52795.2021.00014}
\showDOI{\tempurl}


\bibitem[Han et~al\mbox{.}(2021)]%
        {legodnn2021mobicom}
\bibfield{author}{\bibinfo{person}{Rui Han}, \bibinfo{person}{Qinglong Zhang},
  \bibinfo{person}{Chi~Harold Liu}, \bibinfo{person}{Guoren Wang},
  \bibinfo{person}{Jian Tang}, {and} \bibinfo{person}{Lydia~Y. Chen}.}
  \bibinfo{year}{2021}\natexlab{}.
\newblock \showarticletitle{LegoDNN: block-grained scaling of deep neural
  networks for mobile vision}. In \bibinfo{booktitle}{\emph{{ACM} MobiCom '21:
  The 27th Annual International Conference on Mobile Computing and Networking,
  New Orleans, Louisiana, USA, October 25-29, 2021}}.
  \bibinfo{publisher}{{ACM}}, \bibinfo{pages}{406--419}.
\newblock
\urldef\tempurl%
\url{https://doi.org/10.1145/3447993.3483249}
\showDOI{\tempurl}


\bibitem[He et~al\mbox{.}(2023)]%
        {mtz2023tmc}
\bibfield{author}{\bibinfo{person}{Xiaoxi He}, \bibinfo{person}{Xu Wang},
  \bibinfo{person}{Zimu Zhou}, \bibinfo{person}{Jiahang Wu},
  \bibinfo{person}{Zheng Yang}, {and} \bibinfo{person}{Lothar Thiele}.}
  \bibinfo{year}{2023}\natexlab{}.
\newblock \showarticletitle{On-Device Deep Multi-Task Inference via Multi-Task
  Zipping}.
\newblock \bibinfo{journal}{\emph{{IEEE} Trans. Mob. Comput.}}
  \bibinfo{volume}{22}, \bibinfo{number}{5} (\bibinfo{year}{2023}),
  \bibinfo{pages}{2878--2891}.
\newblock
\urldef\tempurl%
\url{https://doi.org/10.1109/TMC.2021.3124306}
\showDOI{\tempurl}


\bibitem[Ignatov et~al\mbox{.}(2019)]%
        {ai_benchmark2019iccvw}
\bibfield{author}{\bibinfo{person}{Andrey Ignatov} {et~al\mbox{.}}}
  \bibinfo{year}{2019}\natexlab{}.
\newblock \showarticletitle{{AI Benchmark: All About Deep Learning on
  Smartphones in 2019}}. In \bibinfo{booktitle}{\emph{{IEEE/CVF} International
  Conference on Computer Vision Workshops (ICCVW)}}.
\newblock


\bibitem[Iqbal et~al\mbox{.}(2023)]%
        {flexibo2023jair}
\bibfield{author}{\bibinfo{person}{Md~Shahriar Iqbal}, \bibinfo{person}{Jianhai
  Su}, \bibinfo{person}{Lars Kotthoff}, {and} \bibinfo{person}{Pooyan
  Jamshidi}.} \bibinfo{year}{2023}\natexlab{}.
\newblock \showarticletitle{FlexiBO: {A} Decoupled Cost-Aware Multi-Objective
  Optimization Approach for Deep Neural Networks}.
\newblock \bibinfo{journal}{\emph{J. Artif. Intell. Res.}}
  \bibinfo{volume}{77} (\bibinfo{year}{2023}), \bibinfo{pages}{645--682}.
\newblock
\urldef\tempurl%
\url{https://doi.org/10.1613/JAIR.1.14139}
\showDOI{\tempurl}


\bibitem[Jeong et~al\mbox{.}(2022)]%
        {band2022mobisys}
\bibfield{author}{\bibinfo{person}{Joo~Seong Jeong}, \bibinfo{person}{Jingyu
  Lee}, \bibinfo{person}{Donghyun Kim}, \bibinfo{person}{Changmin Jeon},
  \bibinfo{person}{Changjin Jeong}, \bibinfo{person}{Youngki Lee}, {and}
  \bibinfo{person}{Byung{-}Gon Chun}.} \bibinfo{year}{2022}\natexlab{}.
\newblock \showarticletitle{Band: coordinated multi-DNN inference on
  heterogeneous mobile processors}. In \bibinfo{booktitle}{\emph{MobiSys '22:
  The 20th Annual International Conference on Mobile Systems, Applications and
  Services, Portland, Oregon, 27 June 2022 - 1 July 2022}},
  \bibfield{editor}{\bibinfo{person}{Nirupama Bulusu}, \bibinfo{person}{Ehsan
  Aryafar}, \bibinfo{person}{Aruna Balasubramanian}, {and}
  \bibinfo{person}{Junehwa Song}} (Eds.). \bibinfo{publisher}{{ACM}},
  \bibinfo{pages}{235--247}.
\newblock
\urldef\tempurl%
\url{https://doi.org/10.1145/3498361.3538948}
\showDOI{\tempurl}


\bibitem[Jia et~al\mbox{.}(2022)]%
        {codl2022mobisys}
\bibfield{author}{\bibinfo{person}{Fucheng Jia}, \bibinfo{person}{Deyu Zhang},
  \bibinfo{person}{Ting Cao}, \bibinfo{person}{Shiqi Jiang},
  \bibinfo{person}{Yunxin Liu}, \bibinfo{person}{Ju Ren}, {and}
  \bibinfo{person}{Yaoxue Zhang}.} \bibinfo{year}{2022}\natexlab{}.
\newblock \showarticletitle{CoDL: efficient {CPU-GPU} co-execution for deep
  learning inference on mobile devices}. In \bibinfo{booktitle}{\emph{MobiSys
  '22: The 20th Annual International Conference on Mobile Systems, Applications
  and Services, Portland, Oregon, 27 June 2022 - 1 July 2022}},
  \bibfield{editor}{\bibinfo{person}{Nirupama Bulusu}, \bibinfo{person}{Ehsan
  Aryafar}, \bibinfo{person}{Aruna Balasubramanian}, {and}
  \bibinfo{person}{Junehwa Song}} (Eds.). \bibinfo{publisher}{{ACM}},
  \bibinfo{pages}{209--221}.
\newblock
\urldef\tempurl%
\url{https://doi.org/10.1145/3498361.3538932}
\showDOI{\tempurl}


\bibitem[Karatzas and Anagnostopoulos(2023)]%
        {omniboost2023dac}
\bibfield{author}{\bibinfo{person}{Andreas Karatzas} {and}
  \bibinfo{person}{Iraklis Anagnostopoulos}.} \bibinfo{year}{2023}\natexlab{}.
\newblock \showarticletitle{OmniBoost: Boosting Throughput of Heterogeneous
  Embedded Devices under Multi-DNN Workload}.
\newblock \bibinfo{journal}{\emph{CoRR}}  \bibinfo{volume}{abs/2307.03290}
  (\bibinfo{year}{2023}).
\newblock
\urldef\tempurl%
\url{https://doi.org/10.48550/arXiv.2307.03290}
\showDOI{\tempurl}
\showeprint[arXiv]{2307.03290}


\bibitem[Kim and Ha(2023)]%
        {kim2023tc}
\bibfield{author}{\bibinfo{person}{Jangryul Kim} {and} \bibinfo{person}{Soonhoi
  Ha}.} \bibinfo{year}{2023}\natexlab{}.
\newblock \showarticletitle{Energy-Aware Scenario-Based Mapping of Deep
  Learning Applications Onto Heterogeneous Processors Under Real-Time
  Constraints}.
\newblock \bibinfo{journal}{\emph{{IEEE} Trans. Computers}}
  \bibinfo{volume}{72}, \bibinfo{number}{6} (\bibinfo{year}{2023}),
  \bibinfo{pages}{1666--1680}.
\newblock
\urldef\tempurl%
\url{https://doi.org/10.1109/TC.2022.3218991}
\showDOI{\tempurl}


\bibitem[Kim et~al\mbox{.}(2019)]%
        {mulayer2019eurosys}
\bibfield{author}{\bibinfo{person}{Youngsok Kim}, \bibinfo{person}{Joonsung
  Kim}, \bibinfo{person}{Dongju Chae}, \bibinfo{person}{Daehyun Kim}, {and}
  \bibinfo{person}{Jangwoo Kim}.} \bibinfo{year}{2019}\natexlab{}.
\newblock \showarticletitle{$\mu$layer: Low Latency On-Device Inference using
  Cooperative Single-Layer Acceleration and Processor-Friendly Quantization}.
  In \bibinfo{booktitle}{\emph{Proceedings of the Fourteenth EuroSys Conference
  (EuroSys)}}.
\newblock


\bibitem[Kouris et~al\mbox{.}(2023)]%
        {fluidbatching2023iccad}
\bibfield{author}{\bibinfo{person}{Alexandros Kouris},
  \bibinfo{person}{Stylianos~I. Venieris}, \bibinfo{person}{Stefanos
  Laskaridis}, \bibinfo{person}{}, {and} \bibinfo{person}{Nicholas~D. Lane}.}
  \bibinfo{year}{2023}\natexlab{}.
\newblock \showarticletitle{{Fluid Batching: Exit-Aware Preemptive Serving of
  Early-Exit Neural Networks on Edge NPUs}}. In
  \bibinfo{booktitle}{\emph{International Conference on Computer-Aided Design
  (ICCAD)}}.
\newblock


\bibitem[Kundu et~al\mbox{.}(2023)]%
        {tofa2023edge}
\bibfield{author}{\bibinfo{person}{Achintya Kundu}, \bibinfo{person}{Laura
  Wynter}, \bibinfo{person}{Rhui~Dih Lee}, {and} \bibinfo{person}{Luis Angel~D.
  Bathen}.} \bibinfo{year}{2023}\natexlab{}.
\newblock \showarticletitle{Transfer-Once-For-All: {AI} Model Optimization for
  Edge}. In \bibinfo{booktitle}{\emph{{IEEE} International Conference on Edge
  Computing and Communications, {EDGE} 2023, Chicago, IL, USA, July 2-8,
  2023}}, \bibfield{editor}{\bibinfo{person}{Claudio~A. Ardagna},
  \bibinfo{person}{Feras~M. Awaysheh}, \bibinfo{person}{Hongyi Bian},
  \bibinfo{person}{Carl~K. Chang}, \bibinfo{person}{Rong~N. Chang},
  \bibinfo{person}{Fl{\'{a}}via~Coimbra Delicato}, \bibinfo{person}{Nirmit
  Desai}, \bibinfo{person}{Jing Fan}, \bibinfo{person}{Geoffrey~C. Fox},
  \bibinfo{person}{Andrzej Goscinski}, \bibinfo{person}{Zhi Jin},
  \bibinfo{person}{Anna Kobusinska}, {and} \bibinfo{person}{Omer~F. Rana}}
  (Eds.). \bibinfo{publisher}{{IEEE}}, \bibinfo{pages}{26--35}.
\newblock
\urldef\tempurl%
\url{https://doi.org/10.1109/EDGE60047.2023.00017}
\showDOI{\tempurl}


\bibitem[K{\"{u}}t{\"{u}}k{\c{c}}{\"{u}} et~al\mbox{.}(2022)]%
        {kutukcu2022tecs}
\bibfield{author}{\bibinfo{person}{Basar K{\"{u}}t{\"{u}}k{\c{c}}{\"{u}}},
  \bibinfo{person}{Sabur Baidya}, \bibinfo{person}{Anand Raghunathan}, {and}
  \bibinfo{person}{Sujit Dey}.} \bibinfo{year}{2022}\natexlab{}.
\newblock \showarticletitle{Contention Grading and Adaptive Model Selection for
  Machine Vision in Embedded Systems}.
\newblock \bibinfo{journal}{\emph{{ACM} Trans. Embed. Comput. Syst.}}
  \bibinfo{volume}{21}, \bibinfo{number}{5} (\bibinfo{year}{2022}),
  \bibinfo{pages}{55:1--55:29}.
\newblock
\urldef\tempurl%
\url{https://doi.org/10.1145/3520134}
\showDOI{\tempurl}


\bibitem[Kwon et~al\mbox{.}(2023)]%
        {pagedattention2023sosp}
\bibfield{author}{\bibinfo{person}{Woosuk Kwon}, \bibinfo{person}{Zhuohan Li},
  \bibinfo{person}{Siyuan Zhuang}, \bibinfo{person}{Ying Sheng},
  \bibinfo{person}{Lianmin Zheng}, \bibinfo{person}{Cody~Hao Yu},
  \bibinfo{person}{Joseph Gonzalez}, \bibinfo{person}{Hao Zhang}, {and}
  \bibinfo{person}{Ion Stoica}.} \bibinfo{year}{2023}\natexlab{}.
\newblock \showarticletitle{Efficient Memory Management for Large Language
  Model Serving with PagedAttention}. In \bibinfo{booktitle}{\emph{Proceedings
  of the 29th Symposium on Operating Systems Principles (SOSP)}}.
  \bibinfo{pages}{611--626}.
\newblock


\bibitem[Laskaridis et~al\mbox{.}(2024)]%
        {meltingpoint2024arxiv}
\bibfield{author}{\bibinfo{person}{Stefanos Laskaridis},
  \bibinfo{person}{Kleomenis Kateveas}, \bibinfo{person}{Lorenzo Minto}, {and}
  \bibinfo{person}{Hamed Haddadi}.} \bibinfo{year}{2024}\natexlab{}.
\newblock \showarticletitle{MELTing point: Mobile Evaluation of Language
  Transformers}.
\newblock \bibinfo{journal}{\emph{arXiv preprint arXiv:2403.12844}}
  (\bibinfo{year}{2024}).
\newblock


\bibitem[Laskaridis et~al\mbox{.}(2020)]%
        {hapi2020iccad}
\bibfield{author}{\bibinfo{person}{Stefanos Laskaridis},
  \bibinfo{person}{Stylianos~I. Venieris}, \bibinfo{person}{Hyeji Kim}, {and}
  \bibinfo{person}{Nicholas~D. Lane}.} \bibinfo{year}{2020}\natexlab{}.
\newblock \showarticletitle{{HAPI: Hardware-Aware Progressive Inference}}. In
  \bibinfo{booktitle}{\emph{International Conference on Computer-Aided Design
  (ICCAD)}}.
\newblock


\bibitem[Lee et~al\mbox{.}(2022)]%
        {s3nas2022tcad}
\bibfield{author}{\bibinfo{person}{Jaeseong Lee}, \bibinfo{person}{Jungsub
  Rhim}, \bibinfo{person}{Duseok Kang}, {and} \bibinfo{person}{Soonhoi Ha}.}
  \bibinfo{year}{2022}\natexlab{}.
\newblock \showarticletitle{{SNAS:} Fast Hardware-Aware Neural Architecture
  Search Methodology}.
\newblock \bibinfo{journal}{\emph{{IEEE} Trans. Comput. Aided Des. Integr.
  Circuits Syst.}} \bibinfo{volume}{41}, \bibinfo{number}{11}
  (\bibinfo{year}{2022}), \bibinfo{pages}{4826--4836}.
\newblock
\urldef\tempurl%
\url{https://doi.org/10.1109/TCAD.2021.3134843}
\showDOI{\tempurl}


\bibitem[Lee and Nirjon(2020)]%
        {virt_weights2020mobisys}
\bibfield{author}{\bibinfo{person}{Seulki Lee} {and} \bibinfo{person}{Shahriar
  Nirjon}.} \bibinfo{year}{2020}\natexlab{}.
\newblock \showarticletitle{{Fast and Scalable In-Memory Deep Multitask
  Learning via Neural Weight Virtualization}}. In
  \bibinfo{booktitle}{\emph{International Conference on Mobile Systems,
  Applications, and Services (MobiSys)}}.
\newblock


\bibitem[Ling et~al\mbox{.}(2021)]%
        {rtmdl2021sensys}
\bibfield{author}{\bibinfo{person}{Neiwen Ling}, \bibinfo{person}{Kai Wang},
  \bibinfo{person}{Yuze He}, \bibinfo{person}{Guoliang Xing}, {and}
  \bibinfo{person}{Daqi Xie}.} \bibinfo{year}{2021}\natexlab{}.
\newblock \showarticletitle{RT-mDL: Supporting Real-Time Mixed Deep Learning
  Tasks on Edge Platforms}. In \bibinfo{booktitle}{\emph{SenSys '21: The 19th
  {ACM} Conference on Embedded Networked Sensor Systems, Coimbra, Portugal,
  November 15 - 17, 2021}}, \bibfield{editor}{\bibinfo{person}{Jorge~S{\'{a}}
  Silva}, \bibinfo{person}{Fernando Boavida}, \bibinfo{person}{Andr{\'{e}}
  Rodrigues}, \bibinfo{person}{Andrew Markham}, {and} \bibinfo{person}{Rong
  Zheng}} (Eds.). \bibinfo{publisher}{{ACM}}, \bibinfo{pages}{1--14}.
\newblock
\urldef\tempurl%
\url{https://doi.org/10.1145/3485730.3485938}
\showDOI{\tempurl}


\bibitem[Liu et~al\mbox{.}(2021)]%
        {adaspring2021imwut}
\bibfield{author}{\bibinfo{person}{Sicong Liu}, \bibinfo{person}{Bin Guo},
  \bibinfo{person}{Ke Ma}, \bibinfo{person}{Zhiwen Yu}, {and}
  \bibinfo{person}{Junzhao Du}.} \bibinfo{year}{2021}\natexlab{}.
\newblock \showarticletitle{AdaSpring: Context-adaptive and
  Runtime-evolutionary Deep Model Compression for Mobile Applications}.
\newblock \bibinfo{journal}{\emph{Proc. {ACM} Interact. Mob. Wearable
  Ubiquitous Technol.}} \bibinfo{volume}{5}, \bibinfo{number}{1}
  (\bibinfo{year}{2021}), \bibinfo{pages}{24:1--24:22}.
\newblock
\urldef\tempurl%
\url{https://doi.org/10.1145/3448125}
\showDOI{\tempurl}


\bibitem[Mehta and Rastegari(2022)]%
        {mobilevit2022iclr}
\bibfield{author}{\bibinfo{person}{Sachin Mehta} {and}
  \bibinfo{person}{Mohammad Rastegari}.} \bibinfo{year}{2022}\natexlab{}.
\newblock \showarticletitle{MobileViT: Light-weight, General-purpose, and
  Mobile-friendly Vision Transformer}. In \bibinfo{booktitle}{\emph{The Tenth
  International Conference on Learning Representations, {ICLR} 2022, Virtual
  Event, April 25-29, 2022}}. \bibinfo{publisher}{OpenReview.net}.
\newblock
\urldef\tempurl%
\url{https://openreview.net/forum?id=vh-0sUt8HlG}
\showURL{%
\tempurl}


\bibitem[Miret et~al\mbox{.}(2022)]%
        {nemo2022gecco}
\bibfield{author}{\bibinfo{person}{Santiago Miret}, \bibinfo{person}{Vui~Seng
  Chua}, \bibinfo{person}{Mattias Marder}, \bibinfo{person}{Mariano Phiellip},
  \bibinfo{person}{Nilesh Jain}, {and} \bibinfo{person}{Somdeb Majumdar}.}
  \bibinfo{year}{2022}\natexlab{}.
\newblock \showarticletitle{Neuroevolution-enhanced multi-objective
  optimization for mixed-precision quantization}. In
  \bibinfo{booktitle}{\emph{{GECCO} '22: Genetic and Evolutionary Computation
  Conference, Boston, Massachusetts, USA, July 9 - 13, 2022}},
  \bibfield{editor}{\bibinfo{person}{Jonathan~E. Fieldsend} {and}
  \bibinfo{person}{Markus Wagner}} (Eds.). \bibinfo{publisher}{{ACM}},
  \bibinfo{pages}{1057--1065}.
\newblock
\urldef\tempurl%
\url{https://doi.org/10.1145/3512290.3528692}
\showDOI{\tempurl}


\bibitem[Mukherjee et~al\mbox{.}(2021)]%
        {xtremedistiltransformers2021arxiv}
\bibfield{author}{\bibinfo{person}{Subhabrata Mukherjee},
  \bibinfo{person}{Ahmed~Hassan Awadallah}, {and} \bibinfo{person}{Jianfeng
  Gao}.} \bibinfo{year}{2021}\natexlab{}.
\newblock \bibinfo{title}{XtremeDistilTransformers: Task Transfer for
  Task-agnostic Distillation}.
\newblock
\newblock
\showeprint[arxiv]{2106.04563}~[cs.CL]


\bibitem[Panopoulos et~al\mbox{.}(2023)]%
        {mobile_nlp_transformers2023distintsys}
\bibfield{author}{\bibinfo{person}{Ioannis Panopoulos},
  \bibinfo{person}{Sokratis Nikolaidis}, \bibinfo{person}{Stylianos~I.
  Venieris}, {and} \bibinfo{person}{Iakovos~S. Venieris}.}
  \bibinfo{year}{2023}\natexlab{}.
\newblock \showarticletitle{{Exploring the Performance and Efficiency of
  Transformer Models for NLP on Mobile Devices}}. In
  \bibinfo{booktitle}{\emph{IEEE Symposium on Computers and Communications
  (ISCC)}}.
\newblock


\bibitem[Parry et~al\mbox{.}(2021)]%
        {parry2021mlcad}
\bibfield{author}{\bibinfo{person}{Hishan Parry}, \bibinfo{person}{Lei Xun},
  \bibinfo{person}{Amin Sabet}, \bibinfo{person}{Jia Bi},
  \bibinfo{person}{Jonathon~S. Hare}, {and} \bibinfo{person}{Geoff~V.
  Merrett}.} \bibinfo{year}{2021}\natexlab{}.
\newblock \showarticletitle{Dynamic Transformer for Efficient Machine
  Translation on Embedded Devices}. In \bibinfo{booktitle}{\emph{3rd {ACM/IEEE}
  Workshop on Machine Learning for CAD, {MLCAD} 2021, Raleigh, NC, USA, August
  30 - Sept. 3, 2021}}. \bibinfo{publisher}{{IEEE}}, \bibinfo{pages}{1--6}.
\newblock
\urldef\tempurl%
\url{https://doi.org/10.1109/MLCAD52597.2021.9531281}
\showDOI{\tempurl}


\bibitem[Pereira et~al\mbox{.}(2022)]%
        {pereira2022acme}
\bibfield{author}{\bibinfo{person}{Jo{\~a}o Luiz~Junho Pereira},
  \bibinfo{person}{Guilherme~Ant{\^o}nio Oliver},
  \bibinfo{person}{Matheus~Brendon Francisco},
  \bibinfo{person}{Sebasti{\~a}o~Sim{\~o}es Cunha}, {and}
  \bibinfo{person}{Guilherme~Ferreira Gomes}.} \bibinfo{year}{2022}\natexlab{}.
\newblock \showarticletitle{A Review of Multi-objective Optimization: Methods
  and Algorithms in Mechanical Engineering Problems}.
\newblock \bibinfo{journal}{\emph{Archives of Computational Methods in
  Engineering}} \bibinfo{volume}{29}, \bibinfo{number}{4} (\bibinfo{date}{01
  Jun} \bibinfo{year}{2022}), \bibinfo{pages}{2285--2308}.
\newblock
\showISSN{1886-1784}
\urldef\tempurl%
\url{https://doi.org/10.1007/s11831-021-09663-x}
\showDOI{\tempurl}


\bibitem[Quattoni and Torralba(2009)]%
        {mitindoorscenes2009cvpr}
\bibfield{author}{\bibinfo{person}{Ariadna Quattoni} {and}
  \bibinfo{person}{Antonio Torralba}.} \bibinfo{year}{2009}\natexlab{}.
\newblock \showarticletitle{Recognizing indoor scenes}. In
  \bibinfo{booktitle}{\emph{2009 {IEEE} Computer Society Conference on Computer
  Vision and Pattern Recognition {(CVPR} 2009), 20-25 June 2009, Miami,
  Florida, {USA}}}. \bibinfo{publisher}{{IEEE} Computer Society},
  \bibinfo{pages}{413--420}.
\newblock
\urldef\tempurl%
\url{https://doi.org/10.1109/CVPR.2009.5206537}
\showDOI{\tempurl}


\bibitem[Radosavovic et~al\mbox{.}(2020)]%
        {regnet2020cvpr}
\bibfield{author}{\bibinfo{person}{Ilija Radosavovic},
  \bibinfo{person}{Raj~Prateek Kosaraju}, \bibinfo{person}{Ross~B. Girshick},
  \bibinfo{person}{Kaiming He}, {and} \bibinfo{person}{Piotr Doll{\'{a}}r}.}
  \bibinfo{year}{2020}\natexlab{}.
\newblock \showarticletitle{Designing Network Design Spaces}. In
  \bibinfo{booktitle}{\emph{2020 {IEEE/CVF} Conference on Computer Vision and
  Pattern Recognition, {CVPR} 2020, Seattle, WA, USA, June 13-19, 2020}}.
  \bibinfo{publisher}{Computer Vision Foundation / {IEEE}},
  \bibinfo{pages}{10425--10433}.
\newblock
\urldef\tempurl%
\url{https://doi.org/10.1109/CVPR42600.2020.01044}
\showDOI{\tempurl}


\bibitem[Russakovsky et~al\mbox{.}(2015)]%
        {imagenet2015ijcv}
\bibfield{author}{\bibinfo{person}{Olga Russakovsky}, \bibinfo{person}{Jia
  Deng}, \bibinfo{person}{Hao Su}, \bibinfo{person}{Jonathan Krause},
  \bibinfo{person}{Sanjeev Satheesh}, \bibinfo{person}{Sean Ma},
  \bibinfo{person}{Zhiheng Huang}, \bibinfo{person}{Andrej Karpathy},
  \bibinfo{person}{Aditya Khosla}, \bibinfo{person}{Michael Bernstein},
  \bibinfo{person}{Alexander~C. Berg}, {and} \bibinfo{person}{Li Fei-Fei}.}
  \bibinfo{year}{2015}\natexlab{}.
\newblock \showarticletitle{{ImageNet Large Scale Visual Recognition
  Challenge}}.
\newblock \bibinfo{journal}{\emph{International Journal of Computer Vision
  (IJCV)}} \bibinfo{volume}{115}, \bibinfo{number}{3} (\bibinfo{year}{2015}),
  \bibinfo{pages}{211--252}.
\newblock


\bibitem[Sandler et~al\mbox{.}(2018)]%
        {movilenetv22018cvpr}
\bibfield{author}{\bibinfo{person}{Mark Sandler}, \bibinfo{person}{Andrew~G.
  Howard}, \bibinfo{person}{Menglong Zhu}, \bibinfo{person}{Andrey Zhmoginov},
  {and} \bibinfo{person}{Liang{-}Chieh Chen}.} \bibinfo{year}{2018}\natexlab{}.
\newblock \showarticletitle{MobileNetV2: Inverted Residuals and Linear
  Bottlenecks}. In \bibinfo{booktitle}{\emph{2018 {IEEE} Conference on Computer
  Vision and Pattern Recognition, {CVPR} 2018, Salt Lake City, UT, USA, June
  18-22, 2018}}. \bibinfo{publisher}{Computer Vision Foundation / {IEEE}
  Computer Society}, \bibinfo{pages}{4510--4520}.
\newblock


\bibitem[Sanh et~al\mbox{.}(2019)]%
        {sanh2019aaai}
\bibfield{author}{\bibinfo{person}{Victor Sanh}, \bibinfo{person}{Thomas Wolf},
  {and} \bibinfo{person}{Sebastian Ruder}.} \bibinfo{year}{2019}\natexlab{}.
\newblock \showarticletitle{A Hierarchical Multi-Task Approach for Learning
  Embeddings from Semantic Tasks}. In \bibinfo{booktitle}{\emph{The
  Thirty-Third {AAAI} Conference on Artificial Intelligence, {AAAI} 2019, The
  Thirty-First Innovative Applications of Artificial Intelligence Conference,
  {IAAI} 2019, The Ninth {AAAI} Symposium on Educational Advances in Artificial
  Intelligence, {EAAI} 2019, Honolulu, Hawaii, USA, January 27 - February 1,
  2019}}. \bibinfo{publisher}{{AAAI} Press}, \bibinfo{pages}{6949--6956}.
\newblock
\urldef\tempurl%
\url{https://doi.org/10.1609/aaai.v33i01.33016949}
\showDOI{\tempurl}


\bibitem[Saravia et~al\mbox{.}(2018)]%
        {carer2018emnlp}
\bibfield{author}{\bibinfo{person}{Elvis Saravia},
  \bibinfo{person}{Hsien-Chi~Toby Liu}, \bibinfo{person}{Yen-Hao Huang},
  \bibinfo{person}{Junlin Wu}, {and} \bibinfo{person}{Yi-Shin Chen}.}
  \bibinfo{year}{2018}\natexlab{}.
\newblock \showarticletitle{{CARER}: Contextualized Affect Representations for
  Emotion Recognition}. In \bibinfo{booktitle}{\emph{Proceedings of the 2018
  Conference on Empirical Methods in Natural Language Processing}}.
  \bibinfo{publisher}{Association for Computational Linguistics},
  \bibinfo{address}{Brussels, Belgium}, \bibinfo{pages}{3687--3697}.
\newblock


\bibitem[Seo et~al\mbox{.}(2021)]%
        {slo2021taco}
\bibfield{author}{\bibinfo{person}{Wonik Seo}, \bibinfo{person}{Sanghoon Cha},
  \bibinfo{person}{Yeonjae Kim}, \bibinfo{person}{Jaehyuk Huh}, {and}
  \bibinfo{person}{Jongse Park}.} \bibinfo{year}{2021}\natexlab{}.
\newblock \showarticletitle{SLO-Aware Inference Scheduler for Heterogeneous
  Processors in Edge Platforms}.
\newblock \bibinfo{journal}{\emph{{ACM} Trans. Archit. Code Optim.}}
  \bibinfo{volume}{18}, \bibinfo{number}{4} (\bibinfo{year}{2021}),
  \bibinfo{pages}{43:1--43:26}.
\newblock
\urldef\tempurl%
\url{https://doi.org/10.1145/3460352}
\showDOI{\tempurl}


\bibitem[Sharma and Kumar(2022)]%
        {sharma2022acme}
\bibfield{author}{\bibinfo{person}{Shubhkirti Sharma} {and}
  \bibinfo{person}{Vijay Kumar}.} \bibinfo{year}{2022}\natexlab{}.
\newblock \showarticletitle{A Comprehensive Review on Multi-objective
  Optimization Techniques: Past, Present and Future}.
\newblock \bibinfo{journal}{\emph{Archives of Computational Methods in
  Engineering}} \bibinfo{volume}{29}, \bibinfo{number}{7} (\bibinfo{date}{01
  Nov} \bibinfo{year}{2022}), \bibinfo{pages}{5605--5633}.
\newblock
\showISSN{1886-1784}
\urldef\tempurl%
\url{https://doi.org/10.1007/s11831-022-09778-9}
\showDOI{\tempurl}


\bibitem[She et~al\mbox{.}(2023)]%
        {slo2023iwqos}
\bibfield{author}{\bibinfo{person}{Yechao She}, \bibinfo{person}{Minming Li},
  \bibinfo{person}{Yang Jin}, \bibinfo{person}{Meng Xu},
  \bibinfo{person}{Jianping Wang}, {and} \bibinfo{person}{Bin Liu}.}
  \bibinfo{year}{2023}\natexlab{}.
\newblock \showarticletitle{On-demand Edge Inference Scheduling with Accuracy
  and Deadline Guarantee}. In \bibinfo{booktitle}{\emph{31st {IEEE/ACM}
  International Symposium on Quality of Service, IWQoS 2023, Orlando, FL, USA,
  June 19-21, 2023}}. \bibinfo{publisher}{{IEEE}}, \bibinfo{pages}{1--10}.
\newblock
\urldef\tempurl%
\url{https://doi.org/10.1109/IWQOS57198.2023.10188769}
\showDOI{\tempurl}


\bibitem[Singh et~al\mbox{.}(2020)]%
        {dvfs_survey2020ieeedt}
\bibfield{author}{\bibinfo{person}{Amit~Kumar Singh}, \bibinfo{person}{Somdip
  Dey}, \bibinfo{person}{Klaus McDonald-Maier},
  \bibinfo{person}{Karunakar~Reddy Basireddy}, \bibinfo{person}{Geoff~V.
  Merrett}, {and} \bibinfo{person}{Bashir~M. Al-Hashimi}.}
  \bibinfo{year}{2020}\natexlab{}.
\newblock \showarticletitle{{Dynamic Energy and Thermal Management of
  Multi-core Mobile Platforms: A Survey}}.
\newblock \bibinfo{journal}{\emph{IEEE Design \& Test}} \bibinfo{volume}{37},
  \bibinfo{number}{5} (\bibinfo{year}{2020}), \bibinfo{pages}{25--33}.
\newblock


\bibitem[Sun et~al\mbox{.}(2020)]%
        {mobilebert2020acl}
\bibfield{author}{\bibinfo{person}{Zhiqing Sun}, \bibinfo{person}{Hongkun Yu},
  \bibinfo{person}{Xiaodan Song}, \bibinfo{person}{Renjie Liu},
  \bibinfo{person}{Yiming Yang}, {and} \bibinfo{person}{Denny Zhou}.}
  \bibinfo{year}{2020}\natexlab{}.
\newblock \showarticletitle{MobileBERT: a Compact Task-Agnostic {BERT} for
  Resource-Limited Devices}. In \bibinfo{booktitle}{\emph{Proceedings of the
  58th Annual Meeting of the Association for Computational Linguistics, {ACL}
  2020, Online, July 5-10, 2020}}, \bibfield{editor}{\bibinfo{person}{Dan
  Jurafsky}, \bibinfo{person}{Joyce Chai}, \bibinfo{person}{Natalie Schluter},
  {and} \bibinfo{person}{Joel~R. Tetreault}} (Eds.).
  \bibinfo{publisher}{Association for Computational Linguistics},
  \bibinfo{pages}{2158--2170}.
\newblock
\urldef\tempurl%
\url{https://doi.org/10.18653/V1/2020.ACL-MAIN.195}
\showDOI{\tempurl}


\bibitem[Tan et~al\mbox{.}(2019)]%
        {mnasnet2019cvpr}
\bibfield{author}{\bibinfo{person}{Mingxing Tan}, \bibinfo{person}{Bo Chen},
  \bibinfo{person}{Ruoming Pang}, \bibinfo{person}{Vijay Vasudevan},
  \bibinfo{person}{Mark Sandler}, \bibinfo{person}{Andrew Howard}, {and}
  \bibinfo{person}{Quoc~V. Le}.} \bibinfo{year}{2019}\natexlab{}.
\newblock \showarticletitle{MnasNet: Platform-Aware Neural Architecture Search
  for Mobile}. In \bibinfo{booktitle}{\emph{{IEEE} Conference on Computer
  Vision and Pattern Recognition, {CVPR} 2019, Long Beach, CA, USA, June 16-20,
  2019}}. \bibinfo{publisher}{Computer Vision Foundation / {IEEE}},
  \bibinfo{pages}{2820--2828}.
\newblock
\urldef\tempurl%
\url{https://doi.org/10.1109/CVPR.2019.00293}
\showDOI{\tempurl}


\bibitem[Tan and Le(2019)]%
        {efficientnet2019icml}
\bibfield{author}{\bibinfo{person}{Mingxing Tan} {and} \bibinfo{person}{Quoc~V.
  Le}.} \bibinfo{year}{2019}\natexlab{}.
\newblock \showarticletitle{EfficientNet: Rethinking Model Scaling for
  Convolutional Neural Networks}. In \bibinfo{booktitle}{\emph{Proceedings of
  the 36th International Conference on Machine Learning, {ICML} 2019, 9-15 June
  2019, Long Beach, California, {USA}}} \emph{(\bibinfo{series}{Proceedings of
  Machine Learning Research}, Vol.~\bibinfo{volume}{97})},
  \bibfield{editor}{\bibinfo{person}{Kamalika Chaudhuri} {and}
  \bibinfo{person}{Ruslan Salakhutdinov}} (Eds.). \bibinfo{publisher}{{PMLR}},
  \bibinfo{pages}{6105--6114}.
\newblock


\bibitem[Tan and Le(2021)]%
        {efficientnetv22021icml}
\bibfield{author}{\bibinfo{person}{Mingxing Tan} {and} \bibinfo{person}{Quoc~V.
  Le}.} \bibinfo{year}{2021}\natexlab{}.
\newblock \showarticletitle{EfficientNetV2: Smaller Models and Faster
  Training}. In \bibinfo{booktitle}{\emph{Proceedings of the 38th International
  Conference on Machine Learning, {ICML} 2021, 18-24 July 2021, Virtual Event}}
  \emph{(\bibinfo{series}{Proceedings of Machine Learning Research},
  Vol.~\bibinfo{volume}{139})}, \bibfield{editor}{\bibinfo{person}{Marina
  Meila} {and} \bibinfo{person}{Tong Zhang}} (Eds.).
  \bibinfo{publisher}{{PMLR}}, \bibinfo{pages}{10096--10106}.
\newblock
\urldef\tempurl%
\url{http://proceedings.mlr.press/v139/tan21a.html}
\showURL{%
\tempurl}


\bibitem[Touvron et~al\mbox{.}(2023)]%
        {touvron2023llama}
\bibfield{author}{\bibinfo{person}{Hugo Touvron}, \bibinfo{person}{Thibaut
  Lavril}, \bibinfo{person}{Gautier Izacard}, \bibinfo{person}{Xavier
  Martinet}, \bibinfo{person}{Marie-Anne Lachaux},
  \bibinfo{person}{Timoth{\'e}e Lacroix}, \bibinfo{person}{Baptiste
  Rozi{\`e}re}, \bibinfo{person}{Naman Goyal}, \bibinfo{person}{Eric Hambro},
  \bibinfo{person}{Faisal Azhar}, {et~al\mbox{.}}}
  \bibinfo{year}{2023}\natexlab{}.
\newblock \showarticletitle{LLaMa: Open and Efficient Foundation Language
  Models}.
\newblock \bibinfo{journal}{\emph{arXiv preprint arXiv:2302.13971}}
  (\bibinfo{year}{2023}).
\newblock


\bibitem[Venieris et~al\mbox{.}(2023)]%
        {multidnn_accel2023mc}
\bibfield{author}{\bibinfo{person}{Stylianos~I. Venieris},
  \bibinfo{person}{Christos-Savvas Bouganis}, {and}
  \bibinfo{person}{Nicholas~D. Lane}.} \bibinfo{year}{2023}\natexlab{}.
\newblock \showarticletitle{{Multiple-Deep Neural Network Accelerators for
  Next-Generation Artificial Intelligence Systems}}.
\newblock \bibinfo{journal}{\emph{Computer}} \bibinfo{volume}{56},
  \bibinfo{number}{3} (\bibinfo{year}{2023}), \bibinfo{pages}{70--79}.
\newblock


\bibitem[Venieris et~al\mbox{.}(2021)]%
        {oodin2021smartcomp}
\bibfield{author}{\bibinfo{person}{Stylianos~I. Venieris},
  \bibinfo{person}{Ioannis Panopoulos}, {and} \bibinfo{person}{Iakovos~S.
  Venieris}.} \bibinfo{year}{2021}\natexlab{}.
\newblock \showarticletitle{{OODIn: An Optimised On-Device Inference Framework
  for Heterogeneous Mobile Devices}}. In \bibinfo{booktitle}{\emph{IEEE
  International Conference on Smart Computing (SMARTCOMP)}}.
\newblock


\bibitem[Wang et~al\mbox{.}(2019)]%
        {wang2019high}
\bibfield{author}{\bibinfo{person}{Siqi Wang}, \bibinfo{person}{Gayathri
  Ananthanarayanan}, \bibinfo{person}{Yifan Zeng}, \bibinfo{person}{Neeraj
  Goel}, \bibinfo{person}{Anuj Pathania}, {and} \bibinfo{person}{Tulika
  Mitra}.} \bibinfo{year}{2019}\natexlab{}.
\newblock \showarticletitle{High-Throughput CNN Inference on Embedded ARM
  big.LITTLE Multi-Core Processors}.
\newblock \bibinfo{journal}{\emph{IEEE Transactions on Computer-Aided Design of
  Integrated Circuits and Systems (TCAD)}} \bibinfo{volume}{39},
  \bibinfo{number}{10} (\bibinfo{year}{2019}), \bibinfo{pages}{2254--2267}.
\newblock


\bibitem[Wang et~al\mbox{.}(2021)]%
        {moocompression2021cim}
\bibfield{author}{\bibinfo{person}{Zhehui Wang}, \bibinfo{person}{Tao Luo},
  \bibinfo{person}{Miqing Li}, \bibinfo{person}{Joey~Tianyi Zhou},
  \bibinfo{person}{Rick Siow~Mong Goh}, {and} \bibinfo{person}{Liangli Zhen}.}
  \bibinfo{year}{2021}\natexlab{}.
\newblock \showarticletitle{Evolutionary Multi-Objective Model Compression for
  Deep Neural Networks}.
\newblock \bibinfo{journal}{\emph{{IEEE} Comput. Intell. Mag.}}
  \bibinfo{volume}{16}, \bibinfo{number}{3} (\bibinfo{year}{2021}),
  \bibinfo{pages}{10--21}.
\newblock
\urldef\tempurl%
\url{https://doi.org/10.1109/MCI.2021.3084393}
\showDOI{\tempurl}


\bibitem[Wen et~al\mbox{.}(2023)]%
        {adaptivenet2023corr}
\bibfield{author}{\bibinfo{person}{Hao Wen}, \bibinfo{person}{Yuanchun Li},
  \bibinfo{person}{Zunshuai Zhang}, \bibinfo{person}{Shiqi Jiang},
  \bibinfo{person}{Xiaozhou Ye}, \bibinfo{person}{Ye Ouyang},
  \bibinfo{person}{Ya{-}Qin Zhang}, {and} \bibinfo{person}{Yunxin Liu}.}
  \bibinfo{year}{2023}\natexlab{}.
\newblock \showarticletitle{AdaptiveNet: Post-deployment Neural Architecture
  Adaptation for Diverse Edge Environments}.
\newblock \bibinfo{journal}{\emph{CoRR}}  \bibinfo{volume}{abs/2303.07129}
  (\bibinfo{year}{2023}).
\newblock
\urldef\tempurl%
\url{https://doi.org/10.48550/arXiv.2303.07129}
\showDOI{\tempurl}
\showeprint[arXiv]{2303.07129}


\bibitem[Wu et~al\mbox{.}(2019a)]%
        {fbnet2019cvpr}
\bibfield{author}{\bibinfo{person}{Bichen Wu}, \bibinfo{person}{Xiaoliang Dai},
  \bibinfo{person}{Peizhao Zhang}, \bibinfo{person}{Yanghan Wang},
  \bibinfo{person}{Fei Sun}, \bibinfo{person}{Yiming Wu},
  \bibinfo{person}{Yuandong Tian}, \bibinfo{person}{Peter Vajda},
  \bibinfo{person}{Yangqing Jia}, {and} \bibinfo{person}{Kurt Keutzer}.}
  \bibinfo{year}{2019}\natexlab{a}.
\newblock \showarticletitle{FBNet: Hardware-Aware Efficient ConvNet Design via
  Differentiable Neural Architecture Search}. In
  \bibinfo{booktitle}{\emph{{IEEE} Conference on Computer Vision and Pattern
  Recognition, {CVPR} 2019, Long Beach, CA, USA, June 16-20, 2019}}.
  \bibinfo{publisher}{Computer Vision Foundation / {IEEE}},
  \bibinfo{pages}{10734--10742}.
\newblock
\urldef\tempurl%
\url{https://doi.org/10.1109/CVPR.2019.01099}
\showDOI{\tempurl}


\bibitem[Wu et~al\mbox{.}(2019b)]%
        {fb_edge2019hpca}
\bibfield{author}{\bibinfo{person}{Carole{-}Jean Wu} {et~al\mbox{.}}}
  \bibinfo{year}{2019}\natexlab{b}.
\newblock \showarticletitle{{Machine Learning at Facebook: Understanding
  Inference at the Edge}}. In \bibinfo{booktitle}{\emph{IEEE International
  Symposium on High Performance Computer Architecture (HPCA)}}.
  \bibinfo{pages}{331--344}.
\newblock


\bibitem[Wu et~al\mbox{.}(2021)]%
        {switchflow2021middleware}
\bibfield{author}{\bibinfo{person}{Xiaofeng Wu}, \bibinfo{person}{Jia Rao},
  \bibinfo{person}{Wei Chen}, \bibinfo{person}{Hang Huang},
  \bibinfo{person}{Chris H.~Q. Ding}, {and} \bibinfo{person}{Heng Huang}.}
  \bibinfo{year}{2021}\natexlab{}.
\newblock \showarticletitle{SwitchFlow: preemptive multitasking for deep
  learning}. In \bibinfo{booktitle}{\emph{Middleware '21: 22nd International
  Middleware Conference, Qu{\'{e}}bec City, Canada, December 6 - 10, 2021}},
  \bibfield{editor}{\bibinfo{person}{Kaiwen Zhang},
  \bibinfo{person}{Abdelouahed Gherbi}, \bibinfo{person}{Nalini
  Venkatasubramanian}, {and} \bibinfo{person}{Lu{\'{\i}}s Veiga}} (Eds.).
  \bibinfo{publisher}{{ACM}}, \bibinfo{pages}{146--158}.
\newblock
\urldef\tempurl%
\url{https://doi.org/10.1145/3464298.3493391}
\showDOI{\tempurl}


\bibitem[Xie et~al\mbox{.}(2022)]%
        {scalenet2022eccv}
\bibfield{author}{\bibinfo{person}{Jiyang Xie}, \bibinfo{person}{Xiu Su},
  \bibinfo{person}{Shan You}, \bibinfo{person}{Zhanyu Ma}, \bibinfo{person}{Fei
  Wang}, {and} \bibinfo{person}{Chen Qian}.} \bibinfo{year}{2022}\natexlab{}.
\newblock \showarticletitle{ScaleNet: Searching for the Model to Scale}. In
  \bibinfo{booktitle}{\emph{Computer Vision - {ECCV} 2022 - 17th European
  Conference, Tel Aviv, Israel, October 23-27, 2022, Proceedings, Part {XXI}}}
  \emph{(\bibinfo{series}{Lecture Notes in Computer Science},
  Vol.~\bibinfo{volume}{13681})}, \bibfield{editor}{\bibinfo{person}{Shai
  Avidan}, \bibinfo{person}{Gabriel~J. Brostow}, \bibinfo{person}{Moustapha
  Ciss{\'{e}}}, \bibinfo{person}{Giovanni~Maria Farinella}, {and}
  \bibinfo{person}{Tal Hassner}} (Eds.). \bibinfo{publisher}{Springer},
  \bibinfo{pages}{104--120}.
\newblock
\urldef\tempurl%
\url{https://doi.org/10.1007/978-3-031-19803-8\_7}
\showDOI{\tempurl}


\bibitem[Xu et~al\mbox{.}(2019)]%
        {dlsmartphones2019www}
\bibfield{author}{\bibinfo{person}{Mengwei Xu} {et~al\mbox{.}}}
  \bibinfo{year}{2019}\natexlab{}.
\newblock \showarticletitle{{A First Look at Deep Learning Apps on
  Smartphones}}. In \bibinfo{booktitle}{\emph{WWW}}.
\newblock


\bibitem[Xu et~al\mbox{.}(2023)]%
        {xu2023tmc}
\bibfield{author}{\bibinfo{person}{Zhiyuan Xu}, \bibinfo{person}{Dejun Yang},
  \bibinfo{person}{Chengxiang Yin}, \bibinfo{person}{Jian Tang},
  \bibinfo{person}{Yanzhi Wang}, {and} \bibinfo{person}{Guoliang Xue}.}
  \bibinfo{year}{2023}\natexlab{}.
\newblock \showarticletitle{A Co-Scheduling Framework for {DNN} Models on
  Mobile and Edge Devices With Heterogeneous Hardware}.
\newblock \bibinfo{journal}{\emph{{IEEE} Trans. Mob. Comput.}}
  \bibinfo{volume}{22}, \bibinfo{number}{3} (\bibinfo{year}{2023}),
  \bibinfo{pages}{1275--1288}.
\newblock
\urldef\tempurl%
\url{https://doi.org/10.1109/TMC.2021.3107424}
\showDOI{\tempurl}


\bibitem[Yang et~al\mbox{.}(2024)]%
        {gmorph2024eurosys}
\bibfield{author}{\bibinfo{person}{Qizheng Yang}, \bibinfo{person}{Tianyi
  Yang}, \bibinfo{person}{Mingcan Xiang}, \bibinfo{person}{Lijun Zhang},
  \bibinfo{person}{Haoliang Wang}, \bibinfo{person}{Marco Serafini}, {and}
  \bibinfo{person}{Hui Guan}.} \bibinfo{year}{2024}\natexlab{}.
\newblock \showarticletitle{GMorph: Accelerating Multi-DNN Inference via Model
  Fusion}. In \bibinfo{booktitle}{\emph{Conference on Computer Systems
  (EuroSys)}}.
\newblock


\bibitem[Yang et~al\mbox{.}(2020)]%
        {mutualnet2020eccv}
\bibfield{author}{\bibinfo{person}{Taojiannan Yang}, \bibinfo{person}{Sijie
  Zhu}, \bibinfo{person}{Chen Chen}, \bibinfo{person}{Shen Yan},
  \bibinfo{person}{Mi Zhang}, {and} \bibinfo{person}{Andrew~R. Willis}.}
  \bibinfo{year}{2020}\natexlab{}.
\newblock \showarticletitle{MutualNet: Adaptive ConvNet via Mutual Learning
  from Network Width and Resolution}. In \bibinfo{booktitle}{\emph{Computer
  Vision - {ECCV} 2020 - 16th European Conference, Glasgow, UK, August 23-28,
  2020, Proceedings, Part {I}}} \emph{(\bibinfo{series}{Lecture Notes in
  Computer Science}, Vol.~\bibinfo{volume}{12346})},
  \bibfield{editor}{\bibinfo{person}{Andrea Vedaldi}, \bibinfo{person}{Horst
  Bischof}, \bibinfo{person}{Thomas Brox}, {and} \bibinfo{person}{Jan{-}Michael
  Frahm}} (Eds.). \bibinfo{publisher}{Springer}, \bibinfo{pages}{299--315}.
\newblock
\urldef\tempurl%
\url{https://doi.org/10.1007/978-3-030-58452-8\_18}
\showDOI{\tempurl}


\bibitem[Yi and Lee(2020)]%
        {heimdall2020mobicom}
\bibfield{author}{\bibinfo{person}{Juheon Yi} {and} \bibinfo{person}{Youngki
  Lee}.} \bibinfo{year}{2020}\natexlab{}.
\newblock \showarticletitle{{Heimdall: Mobile GPU Coordination Platform for
  Augmented Reality Applications}}. In \bibinfo{booktitle}{\emph{Annual
  International Conference on Mobile Computing and Networking (MobiCom)}}.
\newblock


\bibitem[Yu et~al\mbox{.}(2021)]%
        {multitenant_gpu2021iccad}
\bibfield{author}{\bibinfo{person}{Fuxun Yu}, \bibinfo{person}{Shawn Bray},
  \bibinfo{person}{Di Wang}, \bibinfo{person}{Longfei Shangguan},
  \bibinfo{person}{Xulong Tang}, \bibinfo{person}{Chenchen Liu}, {and}
  \bibinfo{person}{Xiang Chen}.} \bibinfo{year}{2021}\natexlab{}.
\newblock \showarticletitle{Automated Runtime-Aware Scheduling for Multi-Tenant
  DNN Inference on GPU}. In \bibinfo{booktitle}{\emph{IEEE/ACM International
  Conference On Computer Aided Design (ICCAD)}}.
\newblock


\bibitem[Yu et~al\mbox{.}(2022b)]%
        {multitenant_gpu_survey2022arxiv}
\bibfield{author}{\bibinfo{person}{Fuxun Yu}, \bibinfo{person}{Di Wang},
  \bibinfo{person}{Longfei Shangguan}, \bibinfo{person}{Minjia Zhang},
  \bibinfo{person}{Chenchen Liu}, {and} \bibinfo{person}{Xiang Chen}.}
  \bibinfo{year}{2022}\natexlab{b}.
\newblock \showarticletitle{A Survey of Multi-Tenant Deep Learning Inference on
  GPU}.
\newblock \bibinfo{journal}{\emph{arXiv preprint arXiv:2203.09040}}
  (\bibinfo{year}{2022}).
\newblock


\bibitem[Yu et~al\mbox{.}(2022a)]%
        {orca2022osdi}
\bibfield{author}{\bibinfo{person}{Gyeong-In Yu}, \bibinfo{person}{Joo~Seong
  Jeong}, \bibinfo{person}{Geon-Woo Kim}, \bibinfo{person}{Soojeong Kim}, {and}
  \bibinfo{person}{Byung-Gon Chun}.} \bibinfo{year}{2022}\natexlab{a}.
\newblock \showarticletitle{ORCA: A Distributed Serving System for
  Transformer-Based Generative Models}. In \bibinfo{booktitle}{\emph{16th
  USENIX Symposium on Operating Systems Design and Implementation (OSDI)}}.
  \bibinfo{pages}{521--538}.
\newblock


\bibitem[Yu and Huang(2019)]%
        {slimmable2019iccv}
\bibfield{author}{\bibinfo{person}{Jiahui Yu} {and} \bibinfo{person}{Thomas~S.
  Huang}.} \bibinfo{year}{2019}\natexlab{}.
\newblock \showarticletitle{Universally Slimmable Networks and Improved
  Training Techniques}. In \bibinfo{booktitle}{\emph{2019 {IEEE/CVF}
  International Conference on Computer Vision, {ICCV} 2019, Seoul, Korea
  (South), October 27 - November 2, 2019}}. \bibinfo{publisher}{{IEEE}},
  \bibinfo{pages}{1803--1811}.
\newblock
\urldef\tempurl%
\url{https://doi.org/10.1109/ICCV.2019.00189}
\showDOI{\tempurl}


\bibitem[Yuan et~al\mbox{.}(2023)]%
        {mlink2023pami}
\bibfield{author}{\bibinfo{person}{Mu Yuan}, \bibinfo{person}{Lan Zhang},
  \bibinfo{person}{Zimu Zheng}, \bibinfo{person}{Yi{-}Nan Zhang}, {and}
  \bibinfo{person}{Xiang{-}Yang Li}.} \bibinfo{year}{2023}\natexlab{}.
\newblock \showarticletitle{MLink: Linking Black-Box Models From Multiple
  Domains for Collaborative Inference}.
\newblock \bibinfo{journal}{\emph{{IEEE} Trans. Pattern Anal. Mach. Intell.}}
  \bibinfo{volume}{45}, \bibinfo{number}{10} (\bibinfo{year}{2023}),
  \bibinfo{pages}{12085--12097}.
\newblock
\urldef\tempurl%
\url{https://doi.org/10.1109/TPAMI.2023.3283780}
\showDOI{\tempurl}


\bibitem[Zhang et~al\mbox{.}(2021)]%
        {nnmeter2021mobisys}
\bibfield{author}{\bibinfo{person}{Li~Lyna Zhang}, \bibinfo{person}{Shihao
  Han}, \bibinfo{person}{Jianyu Wei}, \bibinfo{person}{Ningxin Zheng},
  \bibinfo{person}{Ting Cao}, \bibinfo{person}{Yuqing Yang}, {and}
  \bibinfo{person}{Yunxin Liu}.} \bibinfo{year}{2021}\natexlab{}.
\newblock \showarticletitle{nn-Meter: towards accurate latency prediction of
  deep-learning model inference on diverse edge devices}. In
  \bibinfo{booktitle}{\emph{MobiSys '21: The 19th Annual International
  Conference on Mobile Systems, Applications, and Services, Virtual Event,
  Wisconsin, USA, 24 June - 2 July, 2021}},
  \bibfield{editor}{\bibinfo{person}{Suman Banerjee}, \bibinfo{person}{Luca
  Mottola}, {and} \bibinfo{person}{Xia Zhou}} (Eds.).
  \bibinfo{publisher}{{ACM}}, \bibinfo{pages}{81--93}.
\newblock
\urldef\tempurl%
\url{https://doi.org/10.1145/3458864.3467882}
\showDOI{\tempurl}


\bibitem[Zhang et~al\mbox{.}(2020)]%
        {zhang2020cvpr}
\bibfield{author}{\bibinfo{person}{Li~Lyna Zhang}, \bibinfo{person}{Yuqing
  Yang}, \bibinfo{person}{Yuhang Jiang}, \bibinfo{person}{Wenwu Zhu}, {and}
  \bibinfo{person}{Yunxin Liu}.} \bibinfo{year}{2020}\natexlab{}.
\newblock \showarticletitle{Fast Hardware-Aware Neural Architecture Search}. In
  \bibinfo{booktitle}{\emph{2020 {IEEE/CVF} Conference on Computer Vision and
  Pattern Recognition, {CVPR} Workshops 2020, Seattle, WA, USA, June 14-19,
  2020}}. \bibinfo{publisher}{Computer Vision Foundation / {IEEE}},
  \bibinfo{pages}{2959--2967}.
\newblock
\urldef\tempurl%
\url{https://doi.org/10.1109/CVPRW50498.2020.00354}
\showDOI{\tempurl}


\bibitem[Zhang and Qi(2017)]%
        {zhifei2017cvpr}
\bibfield{author}{\bibinfo{person}{Song~Yang Zhang, Zhifei} {and}
  \bibinfo{person}{Hairong Qi}.} \bibinfo{year}{2017}\natexlab{}.
\newblock \showarticletitle{Age Progression/Regression by Conditional
  Adversarial Autoencoder}. In \bibinfo{booktitle}{\emph{IEEE Conference on
  Computer Vision and Pattern Recognition (CVPR)}}. IEEE.
\newblock


\bibitem[Zhang and Yang(2022)]%
        {zhang2022tkde}
\bibfield{author}{\bibinfo{person}{Yu Zhang} {and} \bibinfo{person}{Qiang
  Yang}.} \bibinfo{year}{2022}\natexlab{}.
\newblock \showarticletitle{A Survey on Multi-Task Learning}.
\newblock \bibinfo{journal}{\emph{{IEEE} Trans. Knowl. Data Eng.}}
  \bibinfo{volume}{34}, \bibinfo{number}{12} (\bibinfo{year}{2022}),
  \bibinfo{pages}{5586--5609}.
\newblock
\urldef\tempurl%
\url{https://doi.org/10.1109/TKDE.2021.3070203}
\showDOI{\tempurl}


\bibitem[Zhang et~al\mbox{.}(2023a)]%
        {bcedge2023corr}
\bibfield{author}{\bibinfo{person}{Ziyang Zhang}, \bibinfo{person}{Huan Li},
  \bibinfo{person}{Yang Zhao}, \bibinfo{person}{Changyao Lin}, {and}
  \bibinfo{person}{Jie Liu}.} \bibinfo{year}{2023}\natexlab{a}.
\newblock \showarticletitle{BCEdge: SLO-Aware {DNN} Inference Services with
  Adaptive Batching on Edge Platforms}.
\newblock \bibinfo{journal}{\emph{CoRR}}  \bibinfo{volume}{abs/2305.01519}
  (\bibinfo{year}{2023}).
\newblock
\urldef\tempurl%
\url{https://doi.org/10.48550/arXiv.2305.01519}
\showDOI{\tempurl}
\showeprint[arXiv]{2305.01519}


\bibitem[Zhang et~al\mbox{.}(2023b)]%
        {octopus2023icsoc}
\bibfield{author}{\bibinfo{person}{Ziyang Zhang}, \bibinfo{person}{Yang Zhao},
  {and} \bibinfo{person}{Jie Liu}.} \bibinfo{year}{2023}\natexlab{b}.
\newblock \showarticletitle{Octopus: SLO-Aware Progressive Inference Serving
  via Deep Reinforcement Learning in Multi-tenant Edge Cluster}. In
  \bibinfo{booktitle}{\emph{International Conference on Service-Oriented
  Computing (ICSOC)}}, \bibfield{editor}{\bibinfo{person}{Flavia Monti},
  \bibinfo{person}{Stefanie Rinderle{-}Ma}, \bibinfo{person}{Antonio~Ruiz
  Cort{\'{e}}s}, \bibinfo{person}{Zibin Zheng}, {and} \bibinfo{person}{Massimo
  Mecella}} (Eds.).
\newblock


\end{thebibliography}
